\documentclass[twoside,11pt]{article}

\usepackage{bbm}
\usepackage{jmlr2e}
\usepackage{amsmath}
\usepackage{xcolor}
\usepackage{bm}
\usepackage{tikz}
\usetikzlibrary{bayesnet}
\usepackage{subcaption}
\usepackage{float}
\usepackage{stackengine}
\usepackage{array}
\usepackage{multirow}
\usepackage{stmaryrd}
\usepackage{pgfplots}
\usepgfplotslibrary{fillbetween}
\usetikzlibrary{patterns}
\usetikzlibrary{shapes.geometric}
\usepackage[ruled,vlined]{algorithm2e}
\usepackage[margin=0.5in]{geometry}

\pgfmathdeclarefunction{gauss}{3}{%
  \pgfmathparse{1/(#3*sqrt(2*pi))*exp(-((#1-#2)^2)/(2*#3^2))}%
}
\pgfmathdeclarefunction{sum_gauss}{5}{%
  \pgfmathparse{1/(3*#3*sqrt(2*pi))*exp(-((#1-#2)^2)/(2*#3^2)) + 2/(3*#5*sqrt(2*pi))*exp(-((#1-#4)^2)/(2*#5^2))}%
}

\newcommand{\kl}[2]{D_{\mathrm{KL}} \left[ \left. \left. #1 \right|\right| #2 \right] }

\definecolor{blue-violet}{rgb}{0.54, 0.17, 0.89}
\definecolor{bluepigment}{rgb}{0.2, 0.2, 0.6}
\makeatletter
\newcommand*\bigcdot{\mathpalette\bigcdot@{.5}}
\newcommand*\bigcdot@[2]{\mathbin{\vcenter{\hbox{\scalebox{#2}{$\m@th#1\bullet$}}}}}
\makeatother
\def\delequal{\mathrel{\ensurestackMath{\stackon[1pt]{=}{\scriptstyle\Delta}}}}
\definecolor{Yellow}{RGB}{211, 176, 15}
\definecolor{Green}{rgb}{0.01, 0.75, 0.24}
\colorlet{Red}{red!50!black}
\colorlet{Blue}{blue!50!black}
\definecolor{Violet}{rgb}{0.56,0.14,0.56}

\newcommand{\heart}{\ensuremath\heartsuit}

\usepackage{setspace}
\jmlrheading{1}{2020}{1-48}{4/00}{10/00}{meila00a}{Théophile Champion and Marek Grze\'s and Howard Bowman}

\ShortHeadings{Branching Time Active Inference}{Champion et al.}
\firstpageno{1}

\begin{document}

\title{Branching Time Active Inference\\
{\small empirical study and complexity class analysis}}

\author{\name Théophile Champion \email tmac3@kent.ac.uk \\
       \addr University of Kent, School of Computing\\
       Canterbury CT2 7NZ, United Kingdom
       \AND
       \name Howard Bowman \email H.Bowman@kent.ac.uk \\
       \addr University of Birmingham, School of Psychology,\\
       Birmingham B15 2TT, United Kingdom\\
       University of Kent, School of Computing\\
       Canterbury CT2 7NZ, United Kingdom
       \AND
       \name Marek Grze\'s \email m.grzes@kent.ac.uk \\
       \addr University of Kent, School of Computing\\
       Canterbury CT2 7NZ, United Kingdom
       }
       
\editor{} 

\maketitle

\begin{abstract}
Active inference is a state-of-the-art framework for modelling the brain that explains a wide range of mechanisms such as habit formation, dopaminergic discharge and curiosity. However, recent implementations suffer from an exponential (space and time) complexity class when computing the prior over all the possible policies up to the time horizon. \citet{DeepAIwithMCMC} used Monte Carlo tree search to address this problem, leading to very good results in two different tasks. Additionally, \citet{AITS_THEORY} proposed a tree search approach based on (temporal) structure learning. This was enabled by the development of a variational message passing approach to active inference \citep{AI_VMP}, which enables compositional construction of Bayesian networks for active inference. However, this message passing tree search approach, which we call branching-time active inference (BTAI), has never been tested empirically. In this paper, we present an experimental study of the approach \citep{AITS_THEORY} in the context of a maze solving agent. In this context, we show that both improved prior preferences and deeper search help mitigate the vulnerability to local minima. Then, we compare BTAI to standard active inference (AcI) on a graph navigation task. We show that for small graphs, both BTAI and AcI successfully solve the task. For larger graphs, AcI exhibits an exponential (space) complexity class, making the approach intractable. However, BTAI explores the space of policies more efficiently, successfully scaling to larger graphs. Then, BTAI was compared to the POMCP algorithm \citep{POMCP} on the frozen lake environment. The experiments suggest that BTAI and the POMCP algorithm accumulate a similar amount of reward. Also, we describe when BTAI receives more rewards than the POMCP agent, and when the opposite is true. Finally, we compared BTAI to the approach of \citet{DeepAIwithMCMC} on the dSprites dataset, and we discussed the pros and cons of each approach.
\end{abstract}


\section{Introduction}

Active inference extends the free energy principle \citep{Friston2010,PITTI2020242} to generative models with actions \citep{FRISTON2016862,AI_TUTO,AI_VMP} and can be regarded as a form of planning as inference \citep{PAI}. This framework has successfully explained a wide range of brain phenomena, such as habit formation \citep{FRISTON2016862}, Bayesian surprise \citep{bayes_surprise}, curiosity \citep{curiosity}, and dopaminergic discharge \citep{dopamine}. It has also been applied to a variety of tasks such as navigation in the Animal AI environment \citep{DeepAIwithMCMC}, robotic control \citep{pezzato2020active,sancaktar2020endtoend,9457162}, multi-vehicle control \citep{BUTZ2019135}, the mountain car problem \citep{catal2020learning}, the game DOOM \citep{CULLEN2018809} and the cart-pole problem \citep{cart_pole}.

Active inference builds on a subfield of Bayesian statistics called variational inference \citep{VI_TUTO}, in which the true posterior is approximated with a variational distribution. This method provides a way to balance the computational cost and accuracy of the posterior distribution. Indeed, the variational approach is only tractable because some statistical dependencies are ignored during the inference process, i.e., the variational distribution is generally assumed to fully factorise, leading to the well known mean-field approximation:
$$Q(X) = \prod_{i} Q(X_i)$$
where $X$ is the set of all hidden variables of the model and $X_i$ represents the i-th hidden variable. \citet{VMP_TUTO} presented a message-based implementation of variational inference, naturally called variational message passing. And more recently, \citet{AI_VMP} rigorously framed active inference as a variational message passing procedure. By combining the Forney factor graph formalism \citep{FFG_TUTO} with the method of \citet{VMP_TUTO}, it becomes possible to create modular implementations of active inference \citep{Simul_AI,DBLP:journals/ijar/CoxLV19} that allows the users to define their own generative models without the burden of deriving the update equations. This paper uses a new software package called Homing Pigeon that implements such a modular approach and the relevant code has been made publicly available on GitHub: \url{https://github.com/ChampiB/Homing-Pigeon}.

Arguably, the major bottleneck for scaling up the active inference framework was the exponential growth of the number of policies. In the reinforcement learning literature, this explosion is frequently handled using Monte Carlo tree search (MCTS) \citep{Go,6145622,MuZero}. MCTS is based on the upper confidence bound for trees (UCT), which originally comes from the multi-armed bandit problem, and trades-off exploration and exploitation during the tree search. In the reinforcement learning litterature, the selection of the node to expand is carried out using the UCT criterion\footnote{This version of UCT comes from \citet{Go}}, which is defined as:
\begin{align}\label{eq:UCT_CRITERIA_ALPHAGO}
UCT(s,a) = q(s,a) + C_{\text{explore}} \frac{P(s,a)}{1+N(s,a)},
\end{align}
where $q(s,a)$ is the value of taking action $a$ in state $s$ (i.e. $q$ here is not the variational posterior), $C_{\text{explore}}$ is the exploration constant that modulates the amount of exploration, $N(s,a)$ is the visit count, and $P(s,a)$ is the prior probability of selecting action $a$ in state $s$. This approach has been applied to active inference in several papers \citep{DeepAIwithMCMC,LargePOMDP}. \citet{DeepAIwithMCMC} chose to modify the original criterion used during the node selection step that returns the node to be expanded. From equation (9) of \citep{DeepAIwithMCMC}, one can see that the UCT formula has been replaced by:
\begin{equation}\label{eq:uct1}
U(s, a) = -\tilde{G}(s, a) + C_{\text{explore}} \,\, \frac{Q(a|s)}{1 + N(s, a)}
\end{equation}
where $U(s, a)$ indicates the utility of selecting action $a$ in state $s$; $N(s, a)$ is the number of times that action $a$ was explored in state $s$; $C_{\text{explore}}$ is an exploration constant equivalent to $C_p$ in the UCT criterion; $Q(a|s)$ is a neural network modelling the posterior distribution over actions, which is trained by minimizing the variational free energy and $\tilde{G}(s, a)$ is the best estimation of the expected free energy (EFE) computed from the following equation:
\begin{align*}
G(\pi, \tau) = &- \mathbb{E}_{Q(\theta|\pi)Q(s_\tau|\theta,\pi)Q(o_\tau|s_\tau,\theta,\pi)}\Big[\ln P(o_\tau|\pi)\Big]\\
&+ \mathbb{E}_{Q(\theta|\pi)}\Big[\mathbb{E}_{Q(o_\tau|\theta,\pi)}H(s_\tau|o_\tau,\pi) - H(s_\tau|\pi)\Big]\\
&+ \mathbb{E}_{Q(\theta|\pi)Q(s_\tau|\theta,\pi)}H(o_\tau|s_\tau,\theta,\pi) - \mathbb{E}_{Q(s_\tau|\pi)}H(o_\tau|s_\tau,\pi),
\end{align*}
using sampling of 3 (out of 4) neural networks\footnote{\citet{DeepAIwithMCMC} used neural networks to model the likelihood mapping $P(o_\tau|s_\tau)$, the transition mapping $P(s_{\tau+1}|s_\tau,a_\tau)$, the posterior over states $Q(s_\tau)$, and the posterior over actions $Q(a_\tau|s_\tau)$} used by the system. Note that $Q(a|s)$ in equation \eqref{eq:uct1} specializes $P(s,a)$ in equation \eqref{eq:UCT_CRITERIA_ALPHAGO}, by providing the probability of selecting action $a$ in state $s$. One can see that $U(s, a)$ in equation \eqref{eq:uct1} has been obtained from $UCT$ in equation \eqref{eq:UCT_CRITERIA_ALPHAGO}, by replacing the average reward by the negative EFE.

More recently, \citet{AITS_THEORY} proposed an online method that frames planning as a form of (temporal) structure learning guided by the expected free energy. This method, called branching-time active inference (BTAI), generalises active inference \citep{FRISTON2016862,AI_VMP,AI_TUTO} and relates to another recently introduced framework for inference and decision making, called sophisticated inference \citep{sophisticated}. Importantly, the generative model of BTAI enables the agent to trade off risk and ambiguity, instead of only seeking for certainty as was the case in \citep{AI_VMP}. In this paper, we provide an empirical study of BTAI, enabling us to explicitly demonstrate that BTAI provides a more scalable realization of planning as inference than active inference.

Section \ref{sec:ai_ts} reviews the BTAI theory, with full details presented in \citep{AITS_THEORY}. Then, Section \ref{sec:main_results} compares BTAI to standard active inference in the context of a graph navigation task both empirically and theoretically. We show that active inference is able to solve small graphs but suffers from an exponential (space and time) complexity class that makes the approach intractable for bigger graphs. In contrast, BTAI is able to search the space of policies efficiently and scale to bigger graphs. Next, Section \ref{sec:results_intuition} presents the challenge of local minima in the context of a maze solving task, and shows how better prior preferences and deeper tree search help to overcome this challenge. Lastly, Section \ref{sec:results_costs} compares two cost functions, $g^{classic}$ and $g^{pcost}$, in two new mazes. In Section \ref{sec:frozen_lake_env}, BTAI was compared to the POMCP algorithm \citep{POMCP} on the frozen lake environment; and the experiments suggest that BTAI and the POMCP algorithm accumulate a similar amount of reward. Also, we describe when BTAI receives more rewards than the POMCP agent, and when the opposite is true. In Section \ref{sec:sprites_env}, BTAI was compared to the approach of \citet{DeepAIwithMCMC} on the dSprites dataset, and we discussed the pros and cons of each approach. Finally, Section \ref{sec:conclusion} concludes this paper, and provides ideas for future research.

\section{Branching Time Active Inference (BTAI)} \label{sec:ai_ts}

In this section, we provide a short review of BTAI, and the reader is referred to \citep{AITS_THEORY} for details. BTAI frames planning as a form of (temporal) structure learning guided by the expected free energy. This form of structure learning should not be confused with representational or parametric structure learning that is currently developped in the literature \citep{structure_learning,FRISTON2016413,BMR}. The idea is to define a generative model that can be expanded dynamically as shown in Figure \ref{fig:AITS}.

The past and present is modelled using a partially observable Markov decision process (POMDP) in which each observation ($O_\tau$) only depends on the state at time $\tau$, and this state ($S_\tau$) only depends on the previous state ($S_{\tau - 1}$) and previous action ($U_{\tau - 1}$). In addition to the POMDP which models the past and present, the future is modelled using a tree-like generative model whose branches are dynamically expanded. Each branch of the tree corresponds to a trajectory of states reached under a specific policy. The branches are expanded following a logic similar to the Monte Carlo tree search algorithm (see below), and the state estimation is performed using variational message passing \citep{VMP_TUTO,AI_VMP,believe}.

{
\definecolor{Green}{rgb}{0.01, 0.75, 0.24}
\colorlet{Red}{red!50!black}
\colorlet{Blue}{blue!50!black}
\renewcommand\fbox{\fcolorbox{white}{white}}
\begin{figure}[H]
	\begin{center}
	\begin{tikzpicture}[square/.style={regular polygon,regular polygon sides=4}]
        \node[latent] (S0) {$S_0$};
        \node[latent, above=of S0, square, scale=0.5, yshift=-1cm] (F0) {$P_{S0}$};
        \node[latent, below=of S0, square, scale=0.5, yshift=1.5cm] (Fi) {$P_{S...}$};
        \node[latent, below=of Fi, yshift=0.5cm] (Si) {$S_{...}$};
        \node[latent, below=of Si, square, scale=0.5, yshift=1.5cm] (Ft) {$P_{St}$};
        \node[latent, below=of Ft, yshift=0.5cm] (St) {$S_t$};
        \node[latent, left=of Fi, xshift=0.5cm] (A0) {$U_0$};
        \node[latent, left=of A0, square, scale=0.5, xshift=1cm] (Fa0) {$P_{U0}$};
        \node[latent, left=of Ft, xshift=0.5cm] (Ai) {$U_{...}$};
        \node[latent, left=of Ai, square, scale=0.5, xshift=1cm] (Fai) {$P_{U...}$};
        \node[latent, right=of S0, square, scale=0.5, xshift=-1.5cm] (Fo0) {$P_{O0}$};
        \node[obs, right=of Fo0, xshift=-0.5cm] (X0) {$O_0$};
        \node[latent, right=of Si, square, scale=0.4, xshift=-1.85cm] (Foi) {$P_{O...}$};
        \node[obs, right=of Foi, xshift=-0.5cm] (Xi) {$O_{...}$};
        \node[latent, right=of St, square, scale=0.5, xshift=-1.5cm] (Fot) {$P_{Ot}$};
        \node[obs, right=of Fot, xshift=-0.5cm] (Xt) {$O_t$};
        \edge {F0} {S0}
        \edge {Fi} {Si}
        \edge {Ft} {St}
        \draw[black] (A0) -- (Fi);
        \draw[black] (S0) -- (Fi);
        \draw[black] (Ai) -- (Ft);
        \draw[black] (Si) -- (Ft);
        \edge {Fa0} {A0}
        \edge {Fai} {Ai}
        \edge {Fo0} {X0}
        \edge {Foi} {Xi}
        \edge {Fot} {Xt}
        \draw[black] (S0) -- (Fo0);
        \draw[black] (Si) -- (Foi);
        \draw[black] (St) -- (Fot);

        \node[latent, below=of St, square, scale=0.35, yshift=1.5cm, xshift=-1.65cm] (F1) {$P_{S(1)}$};
        \node[latent, below=of F1, scale=0.8, yshift=0.5cm, xshift=-0.85cm] (S1) {$S_{(1)}$};
        \node[latent, below=of St, square, scale=0.35, yshift=1.5cm, xshift=1.65cm] (F2) {$P_{S(2)}$};
        \node[latent, below=of F2, scale=0.8, yshift=0.5cm, xshift=0.85cm] (S2) {$S_{(2)}$};
        \node[latent, left=of S1, square, scale=0.35, xshift=1.5cm] (Fo1) {$P_{O(1)}$};
        \node[latent, left=of Fo1, scale=0.8, xshift=0.5cm] (X1) {$O_{(1)}$};
        \node[latent, right=of S2, square, scale=0.35, xshift=-1.5cm] (Fo2) {$P_{O(2)}$};
        \node[latent, right=of Fo2, scale=0.8, xshift=-0.5cm] (X2) {$O_{(2)}$};
        \node[latent, below=of S2, square, scale=0.3, yshift=1.5cm, xshift=1.65cm, draw=lightgray] (F22) {${\color{lightgray}P_{S(22)}}$};
        \node[latent, below=of F22, scale=0.7, yshift=0.5cm, xshift=0.85cm, draw=lightgray] (S22) {${\color{lightgray}S_{(22)}}$};
        \node[latent, below=of S1, square, scale=0.3, yshift=1.5cm, xshift=-1.65cm] (F11) {$P_{S(11)}$};
        \node[latent, below=of F11, scale=0.7, yshift=0.5cm, xshift=-0.85cm] (S11) {$S_{(11)}$};
        \node[latent, below=of S1, square, scale=0.3, yshift=1.5cm, xshift=1.65cm, draw=lightgray] (F12) {${\color{lightgray}P_{S(12)}}$};
        \node[latent, below=of F12, scale=0.7, yshift=0.5cm, xshift=0.85cm, draw=lightgray] (S12) {${\color{lightgray}S_{(12)}}$};
        \node[latent, left=of S11, square, scale=0.3, xshift=1.5cm] (Fo11) {$P_{O(11)}$};
        \node[latent, left=of Fo11, scale=0.7, xshift=0.5cm] (X11) {$O_{(11)}$};

        \draw[black] (St) -- (F1);
        \edge {F1} {S1}
        \draw[black] (St) -- (F2);
        \edge {F2} {S2}
        \draw[black] (S1) -- (Fo1);
        \edge {Fo1} {X1}
        \draw[black] (S11) -- (Fo11);
        \edge {Fo11} {X11}
        \draw[black] (S2) -- (Fo2);
        \edge {Fo2} {X2}
        \draw[lightgray] (S2) -- (F22);
        \edge[lightgray] {F22} {S22}
        \draw[lightgray] (S1) -- (F12);
        \edge[lightgray] {F12} {S12}
        \draw (S1) -- (F11);
        \edge {F11} {S11}
    \end{tikzpicture}
 	\end{center}
\vspace{-0.25cm}
    \caption{
This figure illustrates the expandable generative model allowing planning under active inference. The current time point (the present) is denoted by $t$. All times before $t$ are the past, and after $t$ are the future. States in the future are indexed by multi-index (action sequences), with each digit indicating an action, e.g. $S_{(11)}$. The future is a tree-like generative model whose branches correspond to the policies considered by the agent. The branches can be dynamically expanded during planning and the nodes in light gray represent possible expansions of the current generative model.}
    \label{fig:AITS}
\end{figure}
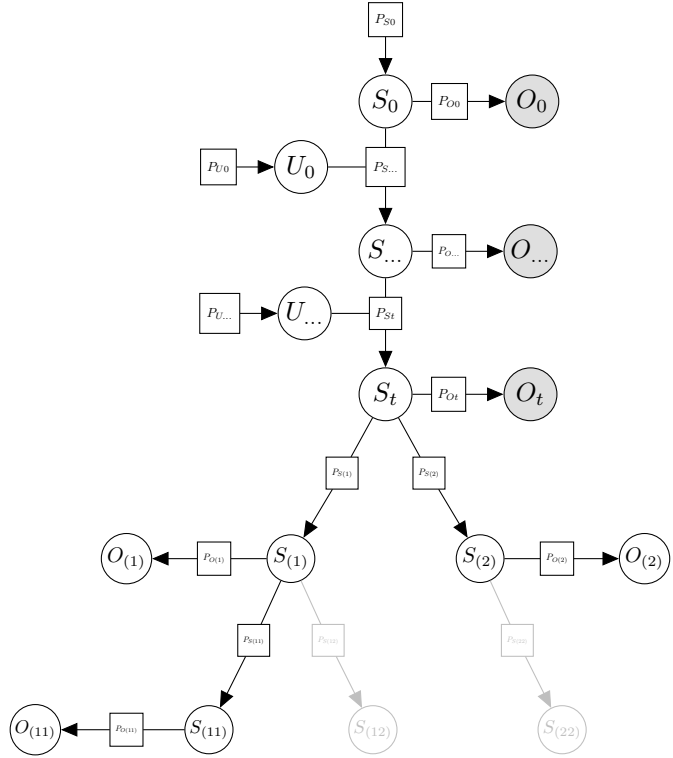
}

At the start of a trial, the model contains only the initial hidden state $S_0$ and the initial observation $O_0$. Then, the agent starts expanding the generative model using an approach inspired by Monte Carlo tree search \citep{6145622}, where the selection of a node is based on expected free energy. More precisely, the node selection is performed recursively from the root until reaching a leaf node. At each level in the recursion the selected node maximises the UCT criterion:
$$UCT_J = \underbrace{- \bar{g}_J}_{\text{exploitation}} +\quad \underbrace{C_p \sqrt{\frac{\ln n}{n_J}}}_{\text{exploration}},$$
where $J$ is a multi-index representing a sequence of actions, $S_J$ is the hidden state reached after performing the actions sequence described by the multi-index $J$, $n$ is the number of times the parent of $S_J$ has been visited, $n_J$ is the number of times the child ($S_J$) was selected, and $\bar{g}_J$ is the average cost received after selecting $S_J$. In what follows, we denote by $J::U$ the the multi-index obtained by adding the action $U$ at the end of the sequence of actions described by the multi-index $J$. Once a leaf node ($S_J$) is selected for expansion, all its children states (i.e., all $S_{J::U}$) are added to the generative model. The future observations (i.e., $O_{J::U}$) associated to those hidden states (i.e., all $S_{J::U}$) are also added to the generative model. Next, the evaluation step estimates the cost of each state-observation pair $(S_{J::U}, O_{J::U})$. In this paper, we consider two kinds of cost. First, the standard expected free energy that trades off risk (over observations) and ambiguity:
$$g^{classic}_J \delequal D_{\mathrm{KL}}[Q(O_J)||V(O_J)]\,\, +\,\, \mathbb{E}_{Q(S_J)}[\text{H}[P(O_J | S_J)]],$$
where $J = I::U$ for an arbitrary action $U$, and $V(O_J)$ is a distribution encoding the prior preferences over observations of the agent, which is generally parameterized by a vector $\bm{C}$ or learnt using a Dirichlet prior \citep{sajid2021exploration}. Second, we also experiment with the following quantity:
$$g^{pcost}_J \delequal \kl{Q(S_J)}{V(S_J)} + \kl{Q(O_J)}{V(O_J)},$$
where $V(S_J)$ is a distribution encoding the prior preferences of the agent over the environment's states. Note that $g^{pcost}_J$ depends on both the risk over observations and the risk over states. The reader is referred to Appendix B for a derivation of $g^{pcost}_J$ from the Free Energy of the Expected Future (FEEF) introduced by \citet{millidge2020expected}. Lastly, the cost of the best action (i.e., the action that produces the smallest cost) is propagated towards the root and used to update the aggregated cost of the ancestors of $S_J$.

Finally, during the planning procedure, the agent needs to perform inference of the future hidden states and observations. This is performed using variational message passing (VMP) on the set of newly expanded nodes, i.e. $\big\{S_{I::U}, O_{I::U} \mid U \in \{1, ..., |U|\}\big\}$, until convergence to a minimum in the free energy landscape. We refer the interested reader to \citep{AI_VMP} for additional information about the derivation of the update equations. Also, since this paper only considers inference and not learning (i.e. the model does not have Dirichlet priors over the tensors defining the world's contingencies), the generative model is different from the one presented in the theoretical paper \citep{AITS_THEORY}. Therefore, we provide a mathematical description of the generative model, the variational distribution and the belief updates in Appendix A. We summarise our method using the pseudo-code in Algorithm \ref{algo: AITS}.

\begin{algorithm}[H]
\label{algo: AITS}
\SetAlgoLined
 \While{end of trial not reached}{
  sample an observation ($O_t$) from the environment\;
  perform inference using VMP and the newly acquired observation ($O_t$)\;
  \While{maximum planning iteration not reached}{
   select a node to be expanded using the UCT criterion\;
   perform the expansion of the generative model from the selected node\;
   perform inference on the newly expanded nodes using VMP\;
   evaluate the cost of the newly expanded nodes using $g^{classic}_J$ or $g^{pcost}_J$\;
   back-propagate the cost of the nodes through the tree\;
  }
  select an action to be performed\;
  execute the action in the environment\;
 }
 \caption{Branching Time Active Inference}
\end{algorithm}

\section{BTAI vs active inference} \label{sec:main_results}

In this section, we benchmark BTAI against standard active inference as implemented in Statistical Parametric Mapping (SPM), c.f. \citet{SPM} for additional details about SPM. First, we do this in terms of complexity class and then empirically through experiments of increasing difficulty.

\subsection{BTAI vs active inference: Space and Time complexity} \label{ssec:complexity_class}

In this section, we compare our model to the standard model of active inference \citep{FRISTON2016862,AI_TUTO}. In the standard formulation, the implementation needs to store the parameter of the posterior over states $\bm{s}^\pi_\tau$ for each policy and each time step. Therefore, assuming $|U|$ possible actions, $T$ time steps, $|\pi| = |U|^T$ policies, and $|S|$ possible hidden state values, the space complexity class for storing the parameters of the posterior over hidden states is $\mathcal{O}(|\pi| \times T \times |S|)$. This corresponds to the number of parameters that needs to be stored, and it is a problem because $|\pi|$ grows exponentially with the number of time steps. Additionally, performing inference on an exponential number of parameters will lead to an exponential time complexity class.

\begin{figure}[H]
	\begin{center}
	\includegraphics[scale=1]{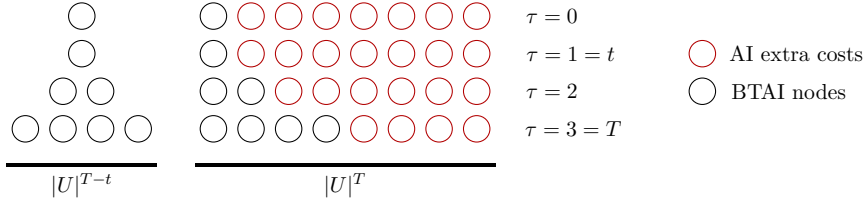}
	\end{center}
\vspace{-0.25cm}
    \caption{
This figure illustrates the difference between AcI and BTAI in terms of space complexity class. The time goes from top to bottom, we assume two actions at each time step, $t$ denotes the current time point, and each circle represents the storage of the $|S|$ parameters required to store a categorical distribution of a hidden state. Black nodes represent the nodes that must be stored in BTAI (under a full expansion of the tree), while the red nodes represent AcI's extra costs of storage. This extra cost comes from the fact that in AcI, one needs to store posterior beliefs for each time step and for each policy, while in BTAI, the tree allows us to compress the representation.}
    \label{fig:complexity_class_theory}
\end{figure}

BTAI solves this problem by allowing only $K$ expansions of the tree. In BTAI, we need to store $|S|$ parameters for each time step in the past and present, and for each expansion, we only need to compute and store the parameters of the posterior over the hidden states corresponding to this expansion. Therefore, the time and space complexity class is $\mathcal{O}([K + t] \times |S|)$, where $t$ is the current time point. This is linear in the number of expansions. Now, the question is how many expansions are required to solve the task? Even if the task requires the tree to be fully expanded, then the complexity class of BTAI would be $\mathcal{O}\big([|U|^{T - t} + t] \times |S|\big)$. Figure \ref{fig:complexity_class_theory} illustrates the difference between AcI and BTAI in terms of the space complexity class, when BTAI performs a full expansion of the tree.

Additionally to the gain afforded by the structure of the tree, most practical applications can be solved by expanding only a small number of nodes \citep{Go,MuZero}, which means that MCTS and BTAI approaches will be even more optimised than in Figure \ref{fig:complexity_class_theory} because most branches will not be expanded.

One could argue that there is a trade off in the nature and extent of the information inferred by classic active inference and branching-time active inference. Specifically, classic active inference exhaustively represents and updates all possible policies, while branching-time active inference will typically only represent a small subset of the possible trajectories. These will typically be the more advantageous paths for the agent to pursue, with the less beneficial paths not represented at all. Indeed, the tree search is based on the expected free energy that favors policies that maximize information gain while realizing the prior preferences of the agent.

Additionally, the inference process can update the system's understanding of past contingencies on the basis of new observations. As a result, the system can obtain more refined information about previous decisions, perhaps re-evaluating the optimality of these past decisions. Because classic active inference represents a larger space of policies, this re-evaluation could apply to more policies.

We also know that humans engage in counterfactual reasoning \citep{rafetseder2013counterfactual}, which, in our planning context, could involve the entertainment and evaluation of alternative (non-selected) sequences of decisions. It may be that, because of the more exhaustive representation of possible trajectories, classic active inference can more efficiently engage in counterfactual reasoning. In contrast, branching-time active inference would require these alternative pasts to be generated ``a fresh" for each counterfactual deliberation. In this sense, one might argue that there is a trade off: branching-time active inference provides considerably more efficient planning to attain current goals, classic active inference provides a more exhaustive assessment of paths not taken. 

\subsection{The deep reward environment}

In this section, we introduce a canonical example of the kind of environment in which BTAI outperforms standard active inference. This environment is called the deep reward environment because the agent needs to navigate a tree like graph, where the graph's nodes correspond to the states of the system, and the agent needs to look deep into the future to diferentiate the favourable path from the traps.

At the beginning of each trial, the agent is placed at the root of the tree that corresponds to the initial state ($S_0$) of the system. From the initial state, the agent can select $m$ actions leading immendiately to an undesirable state, and $n$ actions leading to seemingly pleasant states, for a total of $n + m$ actions. If one of the $m$ undesirable actions is selected, then the agent will enter a bad path, in which (at each time step) $n + m$ actions are available, but all of them produce unpleasant observations. While these $m$ undesirable actions that lead directly to terrible states should be straightforward to avoid for any reasonable agent, the $n$ seemingly favourable actions present an additional challenge. Indeed, only one of those $n$ actions will be beneficial to the agent in the long run, and all the others are long-term traps.

We let $L_k$ with $k \in \{1, ..., n\}$ be the length of the $k$-th seemingly good path. Once the agent is engaged on the $k$-th path, there are still $n + m$ actions available, but only one of them keeps the agent on the right track. All the other actions will produce unpleasant observations, i.e., the agent will enter a bad path. This process will continue until the agent reaches the end of the $k$-th path, which is determined by the path's length $L_k$. If the $k$-th path was the longest of the $n$ seemingly good paths, then the agent will from now on only receive pleasant observations independently of the action performed. If the $k$-th path was not the longest path, then independently of the action performed, the agent will receive painful observations, i.e., the trap is revealed.

To summarize, at the beginning of each trial, the agent is prompted with $n$ seemingly good paths and $m$ obviously bad paths. Only the longest of the seemingly pleasant paths will be beneficial in the long term, the other are traps, which will ultimately lead the agent to an undesirable state. Figure \ref{fig:graph_env} illustrates this environment. Also in theory, this task does not have any terminal states, and the agent will keep taking actions forever. However, in practice, each trial is stopped after a fixed number of action-perception cycles.

{
\definecolor{Green}{rgb}{0.01, 0.75, 0.24}
\colorlet{Red}{red!50!black}
\colorlet{Blue}{blue!50!black}
\renewcommand\fbox{\fcolorbox{white}{white}}
\begin{figure}[H]
	\begin{center}
	\begin{tikzpicture}[square/.style={regular polygon,regular polygon sides=4}]
        \node[latent,fill=white!60!green] (S0) {$S_0$};
        \node[latent,fill=white!60!red,below=of S0,xshift=-4cm] (S1) {$S_b$};
        \node[latent,fill=white!60!red,right=of S1] (Si) {$...$};
        \node[latent,fill=white!60!red,right=of Si] (Sm) {$S_b$};

		\draw [
    		thick,
    		decoration={brace, mirror, raise=0.5cm},
    		decorate
    	] (S1.west) -- (Sm.east) 
    	node [pos=0.5,anchor=north,yshift=-0.55cm] {$m$ bad paths}; 

        \edge {S0} {S1};
        \edge {S0} {Si};
        \edge {S0} {Sm};

        \node[latent,fill=white!60!green,right=of Sm] (S11) {$S_1^{1}$};
        \node[latent,fill=white!60!green,right=of S11] (S1i) {$S_1^{...}$};
        \node[latent,fill=white!60!green,right=of S1i] (S1n) {$S_1^{n}$};

        \edge {S0} {S11};
        \edge {S0} {S1i};
        \edge {S0} {S1n};
        
        \node[latent,fill=white!60!green,below=of S11] (S21) {$S_2^{1}$};
        \node[latent,fill=white!60!red,left=of S21,xshift=0.5cm]   (S21b1) {$S_b$};
        \node[latent,fill=white!60!red,left=of S21b1,xshift=0.5cm] (S21bi) {$...$};
        \node[latent,fill=white!60!red,left=of S21bi,xshift=0.5cm] (S21bl) {$S_b$};

		\draw [
    		thick,
    		decoration={brace, raise=0.5cm},
    		decorate
    	] (S21b1.east) -- (S21bl.west) 
    	node [pos=0.5,anchor=north,yshift=-0.55cm] {$m + n - 1$ bad paths}; 

        \edge {S11} {S21};
        \edge {S11} {S21b1};
        \edge {S11} {S21bi};
        \edge {S11} {S21bl};

        \node[latent,fill=white,draw=white,below=of S1i,yshift=1cm] (S21bi) {$\bm{\vdots}$};

        \node[latent,fill=white!60!green,below=of S21] (Sgi) {$...$};
        \node[latent,fill=white!60!green,left=of Sgi,xshift=0.6cm]   (Sg1) {$S_g$};
        \node[latent,fill=white!60!green,right=of Sgi,xshift=-0.6cm] (Sgl) {$S_g$};

        \edge {S21} {Sg1};
        \edge {S21} {Sgi};
        \edge {S21} {Sgl};

		\draw [
    		thick,
    		decoration={mirror, brace, raise=0.5cm},
    		decorate
    	] (Sg1.west) -- (Sgl.east) 
    	node [pos=0.5,anchor=north,yshift=-0.55cm] {$m + n$ good paths}; 

        \node[latent,fill=white!60!red,below=of S1n] (Snbi) {$...$};
        \node[latent,fill=white!60!red,left=of Snbi,xshift=0.6cm]   (Snb1) {$S_b$};
        \node[latent,fill=white!60!red,right=of Snbi,xshift=-0.6cm] (Snbl) {$S_b$};

        \edge {S1n} {Snbi};
        \edge {S1n} {Snb1};
        \edge {S1n} {Snbl};

		\draw [
    		thick,
    		decoration={mirror, brace, raise=0.5cm},
    		decorate
    	] (Snb1.west) -- (Snbl.east) 
    	node [pos=0.5,anchor=north,yshift=-0.55cm] {$m + n$ bad paths}; 

    \end{tikzpicture}
 	\end{center}
\vspace{-0.25cm}
    \caption{
This figure illustrates a type of environment in which BTAI will outperform standard active inference. Typically, this corresponds to environments in which there are only a small number of good actions. In such environments, BTAI can safely discard a large part of the tree, and speed up the search without impacting performance. Note, $S_0$ represents the initial state, $S_b$ represents a bad state, $S_g$ represents a good state, and $S^i_j$ is the $j$-th state of the $i$-th seemingly good path.
The above picture assumes that the longest path (which is beneficial in the long-term) is the path starting with the state $S^1_1$. Its length ($L_1$) is equal to two because after performing two actions (i.e., the one leading to $S^1_1$ and the one leading to $S^1_2$), the agent is certain to receive pleasant observations. Importantly, any other (seemingly) good path starting with a state $S^i_1$ with $i \in \{2, ..., n\}$ will turn out to be a trap. A trap is simply a state from which all actions lead to an undesirable state ($S_b$), e.g., $S^n_1$ is a trap. Note, at each time point, the agent must pick from the $m + n$ possible actions, e.g, when reaching $S^1_1$ there is only one action keeping the agent on the right track, but all the other actions (i.e., $m + n - 1$ actions) lead to a bad state.}
    \label{fig:graph_env}
\end{figure}
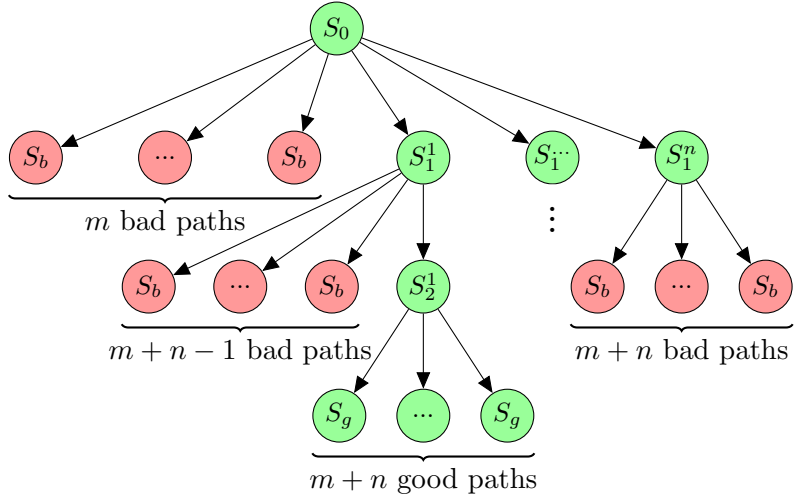
}

\subsubsection{The easy, medium and hard deep reward environment}\label{sssec:instances}

In this section, we present three instances of the deep reward environment in increasing order of complexity (i.e., easy, medium, and hard). These instances will then be used to compare BTAI and (standard) active inference. To specify an instance completely, it is sufficient to provide the number of obviously detrimental actions ($m$), the number of seemingly good actions ($n$), and the length of the paths that follow from the seemingly good actions, i.e., $L_k$ for $k \in \{1, ..., n\}$.

All three instances have five obviously detrimental actions ($m=5$) and two seemingly good actions ($n=2$). However, the lengths of the two good paths (i.e., $L_1$ and $L_2$) change from one instance to the other, and the reader is referred to Table \ref{tab:203} for a summary. In all the environments considered, $L_2 > L_1$, therefore the first path is a trap that will lead to an undesirable state, and the second path is the one that should be taken. Also, to identify that the first path is a trap, the agent must be able to plan at least $L_1 + 1$ steps ahead, since before that the two seemingly good paths are identical. Importantly, an agent trying to evaluate all possible policies $L_1 + 1$ steps into the future, will have to store and process: 343 policies for the easy instance, 16,807 policies for the medium instance, and 5,764,801 policies for the hard instance. We conclude this section with Figure \ref{fig:easy_medium_hard_envs} that illustrates the easy instance of the deep reward environment.

\begin{table}[H]
\centering
\begin{tabular}{ |c|c|c|  }
 \hline
 Environment & $L_1$ & $L_2$\\
 \hline
 easy & 2 & 3\\
 \hline
 medium & 4 & 5\\
 \hline
 hard & 7 & 9\\
 \hline
\end{tabular}
\caption{This table presents the three deep reward environments on which experiments will be run.}
\label{tab:203}
\end{table}

{
\definecolor{Green}{rgb}{0.01, 0.75, 0.24}
\colorlet{Red}{red!50!black}
\colorlet{Blue}{blue!50!black}
\renewcommand\fbox{\fcolorbox{white}{white}}
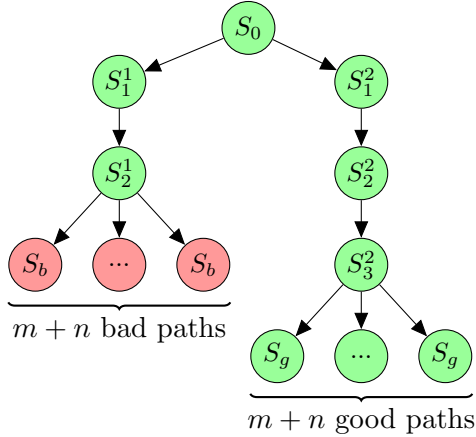
\begin{figure}[H]
	\begin{center}
	\begin{tikzpicture}[square/.style={regular polygon,regular polygon sides=4}]
        \node[latent,fill=white!60!green] (S0) {$S_0$};

        \node[latent,fill=white!60!green,below=of S0,xshift=1.5cm,yshift=1cm] (S11) {$S_1^{2}$};
        \node[latent,fill=white!60!green,left=of S11,xshift=-1.5cm] (S12) {$S_1^{1}$};
        \node[latent,fill=white!60!green,below=of S12,yshift=0.5cm] (S1n) {$S_2^{1}$};

        \edge {S0} {S11};
        \edge {S0} {S12};
        \edge {S12} {S1n};
        
        \node[latent,fill=white!60!green,below=of S11,yshift=0.5cm] (S21) {$S_2^{2}$};

        \edge {S11} {S21};

        \node[latent,fill=white!60!green,below=of S21,yshift=0.5cm] (S31) {$S_3^{2}$};
        \edge {S21} {S31};

        \node[latent,fill=white!60!green,below=of S31,yshift=0.5cm] (Sgi) {$...$};
        \node[latent,fill=white!60!green,left=of Sgi,xshift=0.6cm] (Sg1) {$S_g$};
        \node[latent,fill=white!60!green,right=of Sgi,xshift=-0.6cm] (Sgl) {$S_g$};

        \edge {S31} {Sg1};
        \edge {S31} {Sgi};
        \edge {S31} {Sgl};

		\draw [
    		thick,
    		decoration={mirror, brace, raise=0.5cm},
    		decorate
    	] (Sg1.west) -- (Sgl.east) 
    	node [pos=0.5,anchor=north,yshift=-0.55cm] {$m + n$ good paths}; 

        \node[latent,fill=white!60!red,below=of S1n,yshift=0.5cm] (Snbi) {$...$};
        \node[latent,fill=white!60!red,left=of Snbi,xshift=0.6cm]   (Snb1) {$S_b$};
        \node[latent,fill=white!60!red,right=of Snbi,xshift=-0.6cm] (Snbl) {$S_b$};

        \edge {S1n} {Snbi};
        \edge {S1n} {Snb1};
        \edge {S1n} {Snbl};

		\draw [
    		thick,
    		decoration={mirror, brace, raise=0.5cm},
    		decorate
    	] (Snb1.west) -- (Snbl.east) 
    	node [pos=0.5,anchor=north,yshift=-0.55cm] {$m + n$ bad paths}; 

    \end{tikzpicture}
 	\end{center}
\vspace{-0.25cm}
    \caption{
This figure illustrates the easy instance of the deep reward environment used to compare BTAI and AcI. It contains two seemingly good paths ($n=2$): the first of length two ($L_1=2$) and the second of length three ($L_2=3$). Upon reaching the end of the first (and shortest) path, the agent can only reach undesirable states, i.e., the first path is a trap. In contrast, when reaching the end of the second (and longest) path, the agent can only reach pleasant states, i.e., the second path is beneficial in the long term. Importantly, the entire graph of the easy version contains more than 300 nodes, and is only partially represented. The exhaustive graph is obtained by adding undesirable states ($S_b$) until each node has $n + m$ children, e.g., $S_0$ has $m=5$ unrepresented children and $S_1^1$ has six of them. Finally, the medium and hard versions of the deep reward environment can be obtained from the easy version by lengthening the two seemingly good paths.}
    \label{fig:easy_medium_hard_envs}
\end{figure}
}

\subsection{BTAI vs active inference: Simulations} \label{ssec:BTAI_vs_AI_simul}

In this section, we compare BTAI and active inference on the three instances of the deep reward environment presented in Section \ref{sssec:instances}. The Matlab code running an active inference agent was implemented by modifying the SPM demo called: \url{DEMO_MDP_maze.m}, and is publicly available on GitHub at the following URL: \url{https://github.com/ChampiB/Experiments_AI_TS}, in the file: \url{matlab/graph_navigation.m}.

Table \ref{tab:303.1} shows the result of our simulation in which a standard active inference agent is run on the three deep reward environments presented in Section \ref{sssec:instances}. Since the behaviour of the simulation is deterministic, only one run was executed. If the agent successfully selects the longest path, we report $P(goal) = 1$, otherwise, we report $P(trap) = 1$. Lastly, the simulation was run on a standard laptop with 16GB of RAM, if the agent ran out of memory, then we simply report a ``crash" in the table. As expected, the agent successfully solved the easy and medium environments, for which it was required to plan three and five steps ahead. However, for the hardest version, the agent was supposed to store and process more than five millions policies and the associated beliefs over both: policies and hidden states. This is intractable using only 16GB of RAM and standard active inference runs out of memory because of the exponential (space) complexity class.

\begin{table}[H]
\centering
\begin{tabular}{ |c|c|c|c|c|}
 \hline
 Environment & Policy size & P(goal) & P(trap) & Time (sec)\\
 \hline
 easy & 3 & 1 & 0 & 14.79\\
 medium & 5 & 1 & 0 & 1177.05\\
 hard & 8 & crash & crash & crash \\
 \hline
\end{tabular}
\caption{This table shows that the active inference agent was able to plan three and five time steps ahead to solve the easy and medium deep reward environments. However, because of the exponential space complexity, SPM runs out of memory when trying to plan eight time steps ahead to solve the hardest deep reward environment. The last column reports the time (in seconds) required for running one simulation of the graph environment using SPM.}
\label{tab:303.1}
\end{table}

The C++ code emulating BTAI can be found in the file \url{experiments/main.cpp} of the GitHub repository previously discussed (\url{ChampiB/Experiments_AI_TS}). The hyper-parameters used in the code are described in Appendix D. Since action selection in BTAI is stochastic, we ran 100 simulations. We report the probability of the agent selecting the longest path as: $P(goal) = \frac{\text{number of successes}}{100}$. Simulations where the agent failed to select the proper path are reported as: $P(trap) = \frac{\text{number of failures}}{100}$. We experimented with various numbers of planning iterations, starting with ten iterations and increasing this number by five until the agent was able to solve the task.

Table \ref{tab:303.2} shows the results obtained by BTAI on the three deep reward environments presented in Table \ref{tab:203}, and the hyper-parameter values used in these simulations are reported in Appendix D. As expected, the agent successfully solved the three deep reward environments. Ten planning iterations were required for the easy and medium environments, and twenty for the hardest one. The ability of BTAI to find the best policy among more than five millions policies with only twenty planning iterations is explained by the sparcity of the deep reward environment, i.e., the vast majority of the policies are clearly detrimental to the agent. Note that this sparcity is characteristic of many complex tasks such as chess. For example, a chess player is frequently faced with (chess) positions where twenty to forty legal moves are available, but one move is almost forced, i.e., if not played, the player will almost surely lose the game.

\begin{table}[H]
\centering
\begin{tabular}{ |c|c|c|c|c|  }
 \hline
 Environment & Planning iterations & P(goal) & P(trap) & Time (sec) \\
 \hline
 easy & 10 & 1 & 0 & 0.112 $\pm$ 0.008 \\
 \hline
 medium & 10 & 1 & 0 & 0.193 $\pm$ 0.007 \\
 \hline
 hard & 10 & 0.5 & 0.5 & 0.356 $\pm$ 0.020 \\
  & 15 & 0.49 & 0.51 & 0.536 $\pm$ 0.052 \\
  & 20 & 1 & 0 & 0.836 $\pm$ 0.075 \\
 \hline
\end{tabular}
\caption{This table shows that BTAI was able to solve the three deep reward environments with at most 20 planning iterations. The reported time corresponds to the average runtime of one simulation, and the standard deviation is reported after the symbol $\pm$.}
\label{tab:303.2}
\end{table}

\section{BTAI Empirical Intuition} 

In this section, we study the BTAI agent's behaviour through experiments highlighting its vulnerability to local minimum and ways to mitigate this issue. The goal is to gain some intuition about how the model behaves when: enabling deeper searches, providing better preferences, and using different kind of cost functions to guide the Monte Carlo tree search. The code of those experiments is available on GitHub at the following URL: \url{https://github.com/ChampiB/Experiments_AI_TS}, in the file: \url{experiments/main.cpp}.

\subsection{The maze environment} \label{sec:env}

This section presents the environment in which various simulations will be run. In this environment, the agent can be understood as a rat navigating a maze. Figure \ref{fig:mazes} illustrates the three mazes studied in the following sections. The agent can perform five actions, i.e., UP, DOWN, LEFT, RIGHT and IDLE. The goal is to reach the maze exit from the starting position of the agent. To do so, the agent must move from empty cells to empty cells avoiding walls. If the agent tries to move through a wall, the action becomes equivalent to IDLE. Finally, the observations made by the agent correspond to the Manhattan distance (with the ability to traverse walls) between its current position and the maze exit, i.e.,
$$M(x, y) = \sum_{i = 1}^N |x_i - y_i|,$$
where $M(x, y)$ is the Manhattan distance between $x \in \mathbb{R}^N$ and $y \in \mathbb{R}^N$, $x$ is the position of the agent, $y$ the position of the exit, and in a 2d maze $N=2$. Figure \ref{fig:mazes} (left) illustrates the Manhattan distance received on each cell of a simple maze. Taking maze (A) from Figure \ref{fig:mazes} as an example, if the agent stands on the exit (green square), the observation will be zero or equivalently using one-hot encoding\footnote{A one-hot encoding of a number $n \in \{0, ..., N\}$ means representing $n$ as a vector of size $N + 1$, where the $n$-th element is equal to one and all the other are set to zeros. In this paper, we assume a zero based indexing, i.e., the first element is at index zero.} [1 0 0 0 0 0 0 0 0 0], and if the agent stands at the initial position (red square), the observation will be nine or equivalently [0 0 0 0 0 0 0 0 0 1].

{
\colorlet{Green}{green!70!black}
\colorlet{Red}{red!80!black}
\renewcommand\fbox{\fcolorbox{white}{white}}
\begin{figure}[H]
	\begin{center}
	\begin{tikzpicture}[scale=0.4, every node/.style={scale=0.4}]
	\fill[black] (0,0) rectangle (8,1);
	\fill[black] (0,1) rectangle (1,6);
	\fill[black] (7,1) rectangle (8,6);
	\fill[black] (2,4) rectangle (6,5);
	\fill[black] (5,2) rectangle (6,5);
	\fill[black] (2,2) rectangle (4,3);
	\fill[black] (0,6) rectangle (8,7);
	\fill[Red] (1,1) rectangle (2,2);
	\fill[Green] (6,5) rectangle (7,6);
	\node[scale=2.5] at (6.5,5.5) {0};
	\node[scale=2.5] at (5.5,5.5) {1};
	\node[scale=2.5] at (4.5,5.5) {2};
	\node[scale=2.5] at (3.5,5.5) {3};
	\node[scale=2.5] at (2.5,5.5) {4};
	\node[scale=2.5] at (1.5,5.5) {5};
	\node[scale=2.5] at (1.5,4.5) {6};
	\node[scale=2.5] at (1.5,3.5) {7};
	\node[scale=2.5] at (1.5,2.5) {8};
	\node[scale=2.5] at (2.5,3.5) {6};
	\node[scale=2.5] at (3.5,3.5) {5};
	\node[scale=2.5] at (4.5,3.5) {4};
	\node[scale=2.5] at (4.5,2.5) {5};
	\node[scale=2.5] at (6.5,4.5) {1};
	\node[scale=2.5] at (6.5,3.5) {2};
	\node[scale=2.5] at (6.5,2.5) {3};
	\node[scale=2.5] at (6.5,1.5) {4};
	\node[scale=2.5] at (5.5,1.5) {5};
	\node[scale=2.5] at (4.5,1.5) {6};
	\node[scale=2.5] at (3.5,1.5) {7};
	\node[scale=2.5] at (2.5,1.5) {8};
	\node[scale=2.5] at (1.5,1.5) {9};
	\node[scale=2.5] at (4,-1.5) {(A)};

	\fill[black] (10,1) rectangle (15,2);
	\fill[black] (10,2) rectangle (11,6);
	\fill[black] (14,2) rectangle (15,6);
	\fill[black] (11,5) rectangle (14,6);
	\fill[black] (11,3) rectangle (12,4);
	\fill[black] (13,3) rectangle (14,4);
	\fill[Green] (13,4) rectangle (14,5);
	\fill[Red] (13,2) rectangle (14,3);
	\node[scale=2.5] at (12.5,-0.5) {(B)};

	\fill[black] (17,-1) rectangle (26,0);
	\fill[black] (17,0) rectangle (18,8);
	\fill[black] (23,1) rectangle (24,6);
	\fill[black] (19,3) rectangle (22,4);
	\fill[black] (21,1) rectangle (22,4);
	\fill[black] (19,1) rectangle (20,2);
	\fill[black] (19,5) rectangle (24,6);
	\fill[black] (17,7) rectangle (26,8);
	\fill[black] (25,0) rectangle (26,7);
	\fill[Red] (18,0) rectangle (19,1);
	\fill[Green] (24,6) rectangle (25,7);
	\node[scale=2.5] at (21.5,-2.5) {(C)};

    \end{tikzpicture}
	\end{center}
\vspace{-0.25cm}
    \caption{
This figure illustrates the three mazes used to perform the experiments in the next sections. Black squares correspond to walls, green squares correspond to the maze exit and red squares correspond to the agent starting position. Finally, the numbers displayed on each cell of maze (A) correspond to the Manhattan distance between this cell and the exit.}
    \label{fig:mazes}
\end{figure}
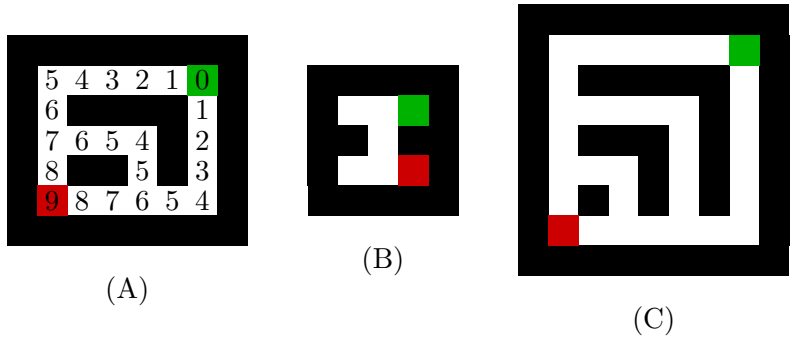
}

\subsection{Overcoming the challenge of local minima}\label{sec:results_intuition}

In this section, we investigate the challenge of local minima and provide two ways of mitigating the issue: improving the prior preferences and using a deeper tree. 

\subsubsection{Prior preferences and local minimum} \label{sec:pp_and_lm}

\begin{figure}[H]
	\begin{center}
	\includegraphics[scale=1]{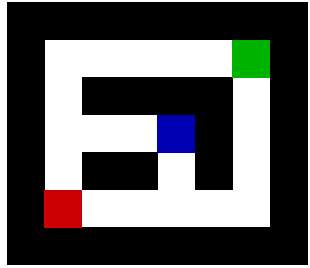}
	\end{center}
\vspace{-0.25cm}
    \caption{
This figure illustrates the notion of local minimum (i.e., the blue cell) in the context of maze (A). Local minima correspond to cells (apart from the exit) for which no adjacent cell has a lower distance to the exit.}
    \label{fig:local_min}
\end{figure}

In this first experiment, the agent was asked to solve maze (B) from Figure \ref{fig:mazes}, which has the property that the start location (red square) is a local minimum. Remember from Section \ref{sec:env} that the agent observes the Manhattan distance between its location and the maze exit. The Manhattan distance naturally creates local minima throughout the mazes, i.e., cells of the maze (apart from the exit) for which no adjacent cell has a lower distance to the exit. An example of such a local minimum is shown as a blue square in Figure \ref{fig:local_min}. The presence of such a local minimum implies that a well behaved agent (i.e., an agent trying to get as close as possible to the exit) might get trapped in those cells for which no adjacent cell has a lower distance to the exit and thus fail to solve the task.

Next, we need to define the prior preferences of the agent. Our framework allows the modeller to define prior preferences over both future observations and future states. However, we start by assuming no preferences over the hidden states, i.e., $V(S_I)$ is uniform. We define the prior preferences over future observations as:
$$\bm{C}_O = \sigma\big(\gamma \bm{v}\big) \text{ with } \bm{v} = \big[\,\, |O| \,\,\,  ... \,\,\, 2 \,\,\, 1 \,\, \big]^T$$
where $|O|$ is the number of possible observations (10 in maze (A) from Figure \ref{fig:mazes}), $\gamma$ is the precision of the prior preferences, and $\sigma(\cdot)$ is the softmax function. The above prior preferences will give high probability to cells close to the exit and will exhibit the local minimum behaviours previously mentioned.

Using these prior preferences, we ran 100 simulations in maze (B) from Figure \ref{fig:mazes}. Each simulation was composed of a maximum of 20 action-perception cycles, and was interrupted when the agent reached the maze exit. Note, the results might vary from simulation to simulation, because the actions performed in the environment are sampled from $\sigma(-\omega \frac{g}{N})$, where $\sigma(\bigcdot)$ is a softmax function, $\omega$ is the precision of action selection, $g$ is a vector whose elements correspond to the cost of the root's children (i.e. the children of $S_t$) and $N$ is a vector whose elements correspond to the number of visits of the root's children.

Table \ref{tab:1} reports the frequency at which the agent reaches the exit. The hyper-parameters values are reported in Appendix D. First, note that with 10 and 15 planning iterations, the agent was unable to leave the initial position (i.e., it is trapped in the local minimum). But as the number of planning iterations is increased, the agent becomes able to foresee the benefits of leaving the local minimum.

\begin{table}[H]
\centering
\begin{tabular}{ |c|c|c|c| }
 \hline
 Planning iterations & P(exit) & P(local) & Time (sec)\\
 \hline
 10 & 0 & 1 & 0.701 $\pm$ 0.022\\
 15 & 0 & 1 & 1.030 $\pm$ 0.070\\
 20 & 1 & 0 & 0.233 $\pm$ 0.018\\
 \hline
\end{tabular}
\caption{This table presents the probability that the agent solves maze (B), and the probability of the agent being stuck into the local minimum. The reported time corresponds to the average runtime of one simulation, and the standard deviation is reported after the symbol $\pm$. Importantly, when the agent reaches the exit of the maze the simulation is interrupted, i.e., the simulation contains less than 20 action-perception cycles. This explains why performing 20 planning iterations is faster (0.233 seconds), than performing 15 planning iterations (1.030 seconds), i.e., the simulations with 15 planning iterations (that fail to solve the maze) contain 20 action-perception cycles while the simulations with 20 planning iterations (that successfully solve the maze) contain less than 20 action-perception cycles.}
\label{tab:1}
\end{table}

\subsubsection{Improving prior preference to avoid local minimum} \label{sec:pp_and_lm_mp}

\begin{figure}[H]
	\begin{center}
	\includegraphics[scale=1]{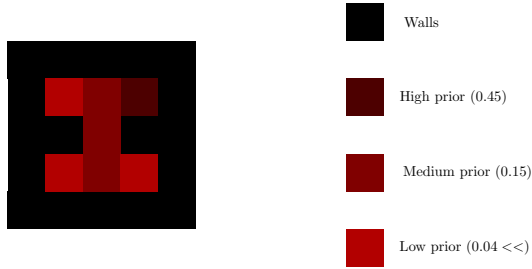}
	\end{center}
\vspace{-0.25cm}
    \caption{
This figure illustrates the new prior preferences of the agent over the future states. Black squares correspond to walls, the darkest red corresponds to high prior preferences (really enjoyable states), the brightest red corresponds to low prior preferences (annoying states) and the last kind of red corresponds to medium prior preferences (boring states).}
    \label{fig:new_prior_pref}
\end{figure}

In this second experiment, we modified the prior preferences of the agent to enable it to avoid local minima. We first change the cost function from the expected free energy $g^{classic}_I$ to the pure cost $g^{pcost}_I$, which allows us to set nontrivial preferences over states (in the previous section, these were set to uniform). Specifically, the prior preferences over hidden states will be of the form:
$$\bm{C}_S = \sigma\big(\gamma \bm{w}\big),$$
where $\gamma$ is the precision over prior preferences, and $\bm{w}$ is set according to Figure \ref{fig:new_prior_pref}. Finally, the prior preferences over future observations remain the same as in the previous section, and once again the hyper-parameters values are reported in Appendix D.

Tables \ref{tab:2} and \ref{tab:2.1} summarize the results of the experiments with and without the use of prior preferences over hidden states, respectively. As expected better prior preferences lead to better performance when less planning iterations are performed. Specifying prior preferences over hidden states requires the modeller to bring additional knowledge to the agent, and might not always be possible. However, when such knowledge is available it can improve the agent's performance. This illustrates the value of the BTAI approach, which enables preferences to be specified for observations, as does active inference, as well as for states.

\begin{table}[H]
\centering
\begin{tabular}{ |c|c|c|c| }
 \hline
 Planning iterations & P(global) & P(local) & Time (sec)\\
 \hline
 10 & 0 & 1 & 0.683 $\pm$ 0.024\\
 \hline
 15 & 0 & 1 & 0.983 $\pm$ 0.030\\
 \hline
 20 & 1 & 0 & 0.217 $\pm$ 0.002\\
 \hline
\end{tabular}
\caption{This table presents the probability that the agent solves maze (B), and the probability of the agent being stuck in the local minimum. In this table, the agent was not equipped with prior preferences over hidden states. The last column reports the (average) execution time required for running one simulation and the associated standard deviation.}
\label{tab:2}
\end{table}

\begin{table}[H]
\centering
\begin{tabular}{ |c|c|c|c| }
 \hline
 Planning iterations & P(global) & P(local) & Time (sec)\\
 \hline
 10 & 0 & 1 & 0.749 $\pm$ 0.045\\
 \hline
 15 & 1 & 0 & 0.181 $\pm$ 0.018\\
 \hline
 20 & 1 & 0 & 0.288 $\pm$ 0.092\\
 \hline
\end{tabular}
\caption{This table presents the probability that the agent solves maze (B), and the probability of the agent being stuck in the local minimum. In this table, the agent was equipped with prior preferences over hidden states. The last column reports the (average) execution time required for running one simulation and the associated standard deviation.}
\label{tab:2.1}
\end{table}

\subsection{Solving more mazes}\label{sec:results_costs}

Up to now, we focused on maze (B) from Figure \ref{fig:mazes} to demonstrate that both improving prior preferences and deepening the tree can help to mitigate the problem of local minima. In this section, we extend our analysis to mazes (A) and (C). Table \ref{tab:103} shows the performance of the BTAI agent in maze (A) when using $g^{classic}_I$ and $g^{pcost}_I$ as cost function. When $g^{pcost}_I$ was used as a cost function, the agent was only equipped with prior preferences over observations (i.e., uniform preferences over hidden states). Table \ref{tab:103.2} shows the results of the same experiments but on maze (C). As usual the hyper-parameters values used for those simulations are given in Appendix D.

Tables \ref{tab:103} and \ref{tab:103.2} seem to indicate that both $g^{classic}_I$ and $g^{pcost}_I$ perform similiarly on the maze environment, and require approximatly the same amount of time to be computed. The similiar performance of $g^{classic}_I$ and $g^{pcost}_I$ may be surprising to the reader. Indeed, $g^{classic}_I$ contains an ambiguity terms, i.e., $\mathbb{E}_{Q(S_J)}[\text{H}[P(O_J | S_J)]]$, which should be helping the agent. In contrast, $g^{pcost}_I$ contains the risk over states with uniform prior preferences over states, i.e., $\kl{Q(S_J)}{V(S_J)}$, which should not be helpful (because of the uniformity of the prior preferences).

However, in the maze envionment the ambiguity of the likelihood mapping $P(O_\tau|S_\tau)$ is identical for each possible hidden state $S_\tau$. Indeed, each state corresponds to a cell, and each cell is at a fix Manhattan distance from the exit. Thus, each state generates with high probability the observation corresponding to the Manhattan distance between the state's cell and the exit; and generates with small probability any other observations. For example, the likelihood mapping of an imaginary maze could be defined as follow:
\begin{align*}
P(O_\tau|S_\tau) = \bm{A} = \begin{bmatrix}
0.05 & 0.05 & 0.9\\
0.05 & 0.9  & 0.05\\
0.9  & 0.05 & 0.05
\end{bmatrix},
\end{align*}
where $P(O_\tau = i|S_\tau = j) = \bm{A}_{ij}$. Importantly, each column of $\bm{A}$ has the same entropy, therefore the agent does not care about which observation is made, i.e., they are all as ambiguous. This is why the ambiguity term is in fact not helpful in the maze environment, and why $g^{classic}_I$ and $g^{pcost}_I$ produce similar performances.

\begin{table}[H]
\centering
\begin{tabular}{ |c|c|c|c|c| }
 \hline
 Planning iterations & P(global) & P(local) & Time (sec) for $g_I^\text{classic}$ & Time (sec) for $g_I^\text{pcost}$ \\
 \hline
 10 & 1 & 0 & 0.310 $\pm$ 0.032 & 0.287 $\pm$ 0.022 \\
 \hline
 15 & 1 & 0 & 0.423 $\pm$ 0.008 & 0.432 $\pm$ 0.011 \\
 \hline
 20 & 1 & 0 & 0.567 $\pm$ 0.026 & 0.579 $\pm$ 0.023 \\
 \hline
\end{tabular}
\caption{This table presents the probability that the agent solves maze (A) from Figure \ref{fig:mazes}, and the probability of the agent falling into the local minimum. Both cost functions $g^{classic}_I$ and $g^{pcost}_I$ lead to the above results in maze (A). The last two columns report the (average) execution time and the associated standard deviation of running one simulation with $g^{classic}_I$ and $g^{pcost}_I$, respectively.}
\label{tab:103}
\end{table}

\begin{table}[H]
\centering
\begin{tabular}{ |c|c|c|c|c| }
 \hline
 Planning iterations & P(global) & P(local) & Time (sec) for $g_I^\text{classic}$ & Time (sec) for $g_I^\text{pcost}$ \\
 \hline
 10 & 1 & 0 & 0.498 $\pm$ 0.053 & 0.460 $\pm$ 0.019 \\
 \hline
 15 & 1 & 0 & 0.696 $\pm$ 0.063 & 0,664 $\pm$ 0.075 \\
 \hline
 20 & 1 & 0 & 0.920 $\pm$ 0.091 & 0.833 $\pm$ 0.038 \\
 \hline
\end{tabular}
\caption{This table presents the probability that the agent solves maze (C), and the probability of the agent falling into the local minimum. Both cost functions $g^{classic}_I$ and $g^{pcost}_I$ lead to the above results in maze (C). The last two columns report the (average) execution time and the associated standard deviation of running one simulation with $g^{classic}_I$ and $g^{pcost}_I$, respectively.}
\label{tab:103.2}
\end{table}

\section{The frozen lake environment}\label{sec:frozen_lake_env}

In this section, we evaluate our agent on the frozen lake environmnent introduced by OpenAI \citep{OpenAI}. The frozen lake environment can be represented as a 2D grid with $r$ rows and $c$ columns. Each cell in the grid is either a frozen surface that can support the agent's weight or a hole on which the agent cannot step without receiving a heavy penalty. One of the cells with a frozen surface contains a frisbee that the agent needs to recover, i.e., this cell is the goal state. For our purpose, each cell is associated with a number describing its location, and the agent observes only its location in the lake. The agent can perform four actions (i.e., UP, DOWN, LEFT, RIGHT) at any point in time. Actions that would lead the agent to leave the lake (through the external boundary), are equivalent to doing nothing and the agent does not move.

{
\colorlet{Green}{green!70!black}
\colorlet{Orange}{orange!70!white}
\colorlet{Red}{red!80!black}
\colorlet{Blue}{blue!50!black}
\colorlet{LightBlue}{blue!50!white}
\renewcommand\fbox{\fcolorbox{white}{white}}
\begin{figure}[H]
	\begin{center}
	\begin{tikzpicture}[scale=0.32, every node/.style={scale=0.32}]
		\node[scale=2.5] at (8,-2) {(a)};

		\fill[black] (0,-1) rectangle (16,0);
		\fill[black] (0,12) rectangle (15,13);
		\fill[black] (0,0) rectangle (1,13);
		\fill[black] (15,0) rectangle (16,13);

		\fill[Orange] (3,10) rectangle (4,11);
		\fill[Orange] (3,6) rectangle (4,7);
		\fill[Orange] (4,7) rectangle (5,8);
		\fill[Orange] (4,0) rectangle (5,1);
		\fill[Orange] (12,0) rectangle (13,1);
		\fill[Orange] (12,3) rectangle (13,4);
		\fill[Orange] (11,5) rectangle (12,6);
		\fill[Orange] (14,10) rectangle (15,11);

		\fill[Blue] (2,10) rectangle (3,11);
		\fill[Blue] (3,9) rectangle (4,10);
		\fill[Blue] (4,10) rectangle (5,12);
		\fill[Blue] (6,9) rectangle (7,10);
		\fill[Blue] (7,10) rectangle (8,11);
		\fill[Blue] (8,11) rectangle (9,12);
		\fill[Blue] (10,11) rectangle (12,12);
		\fill[Blue] (11,10) rectangle (12,11);
		\fill[Blue] (10,9) rectangle (11,10);
		\fill[Blue] (13,10) rectangle (14,11);
		\fill[Blue] (14,10) rectangle (15,8);
		\fill[Blue] (2,8) rectangle (3,6);
		\fill[Blue] (3,6) rectangle (4,5);
		\fill[Blue] (4,7) rectangle (5,6);
		\fill[Blue] (5,8) rectangle (6,7);
		\fill[Blue] (4,5) rectangle (5,4);
		\fill[Blue] (1,5) rectangle (2,4);
		\fill[Blue] (2,3) rectangle (3,1);
		\fill[Blue] (5,3) rectangle (6,2);
		\fill[Blue] (5,0) rectangle (7,1);
		\fill[Blue] (8,3) rectangle (9,2);
		\fill[Blue] (9,2) rectangle (11,1);
		\fill[Blue] (12,3) rectangle (13,2);
		\fill[Blue] (13,4) rectangle (14,3);
		\fill[Blue] (13,0) rectangle (14,1);
		\fill[Blue] (12,7) rectangle (13,5);
		\fill[Blue] (10,4) rectangle (12,5);
		\fill[Blue] (11,7) rectangle (12,8);
		\fill[Blue] (8,8) rectangle (10,7);
		\fill[Blue] (8,6) rectangle (6,5);

		\fill[Red] (1,11) rectangle (2,12);
		\fill[Green] (14,0) rectangle (15,1);

		\node[scale=2.5] at (26,-2) {(b)};

		\fill[black] (18,-1) rectangle (34,0);
		\fill[black] (18,12) rectangle (33,13);
		\fill[black] (18,0) rectangle (19,13);
		\fill[black] (33,0) rectangle (34,13);

		\fill[Blue] (20,10) rectangle (21,11);
		\fill[Blue] (21,9) rectangle (22,10);
		\fill[Blue] (22,10) rectangle (23,12);
		\fill[Blue] (24,9) rectangle (25,10);
		\fill[Blue] (25,10) rectangle (26,11);
		\fill[Blue] (26,11) rectangle (27,12);
		\fill[Blue] (28,11) rectangle (30,12);		
		\fill[Blue] (29,10) rectangle (30,11);
		\fill[Blue] (27,8) rectangle (28,9);
		\fill[Blue] (28,9) rectangle (29,10);
		\fill[Blue] (31,10) rectangle (32,11);
		\fill[Blue] (32,10) rectangle (33,8);
		\fill[Blue] (20,8) rectangle (21,6);
		\fill[Blue] (21,6) rectangle (22,5);
		\fill[Blue] (22,7) rectangle (23,6);
		\fill[Blue] (23,8) rectangle (24,7);
		\fill[Blue] (20,4) rectangle (22,3);
		\fill[Blue] (22,5) rectangle (23,4);
		\fill[Blue] (19,5) rectangle (20,4);
		\fill[Blue] (20,3) rectangle (21,1);
		\fill[Blue] (23,3) rectangle (24,2);
		\fill[Blue] (23,0) rectangle (25,1);
		\fill[Blue] (26,3) rectangle (27,2);
		\fill[Blue] (27,2) rectangle (29,1);
		\fill[Blue] (30,3) rectangle (31,2);
		\fill[Blue] (31,5) rectangle (32,3);
		\fill[Blue] (29,2) rectangle (31,1);
		\fill[Blue] (31,0) rectangle (32,1);
		\fill[Blue] (29,7) rectangle (30,5);
		\fill[Blue] (27,4) rectangle (30,5);
		\fill[Blue] (28,3) rectangle (29,4);
		\fill[Blue] (29,7) rectangle (30,8);
		\fill[Blue] (26,8) rectangle (28,7);
		\fill[Blue] (26,6) rectangle (24,5);

		\fill[Red] (19,11) rectangle (20,12);
		\fill[Green] (32,0) rectangle (33,1);

		\draw [Blue,thick] (19.8,11) -- (19.8,8.8);
		\draw [Blue,thick] (19.8,8.8) -- (24.8,8.8);
		\draw [Blue,thick] (24.8,8.8) -- (24.8,6.8);
		\draw [Blue,thick] (24.8,6.8) -- (28.5,6.8);
		\draw [Blue,thick] (28.5,6.8) -- (28.5,8.5);
		\draw [Blue,thick] (28.5,8.5) -- (30.5,8.5);
		\draw [Blue,thick] (30.5,8.5) -- (30.5,5.6);
		\draw [Blue,thick] (30.5,5.6) -- (32.7,5.6);
		\draw [Blue,thick,-stealth] (32.7,5.6) -- (32.7,1);
		
		\draw [Red,thick] (19.2,11) -- (19.2,8.2);
		\draw [Red,thick] (19.2,8.2) -- (24.2,8.2);
		\draw [Red,thick] (24.2,8.2) -- (24.2,6.2);
		\draw [Red,thick] (24.2,6.2) -- (26.2,6.2);
		\draw [Red,thick] (26.2,6.2) -- (26.2,3.5);
		\draw [Red,thick] (26.2,3.5) -- (25.5,3.5);
		\draw [Red,thick] (25.5,3.5) -- (25.5,0.5);
		\draw [Red,thick,-stealth] (25.5,0.5) -- (30.5,0.5);

		\draw [Green,thick] (19.5,11) -- (19.5,8.5);
		\draw [Green,thick] (19.5,8.5) -- (24.5,8.5);
		\draw [Green,thick] (24.5,8.5) -- (24.5,6.5);
		\draw [Green,thick] (24.5,6.5) -- (26.5,6.5);
		\draw [Green,thick] (26.5,6.5) -- (26.5,3.5);
		\draw [Green,thick] (26.5,3.5) -- (27.5,3.5);
		\draw [Green,thick] (27.5,3.5) -- (27.5,2.5);
		\draw [Green,thick] (27.5,2.5) -- (29.5,2.5);
		\draw [Green,thick] (29.5,2.5) -- (29.5,3.5);
		\draw [Green,thick] (29.5,3.5) -- (30.5,3.5);
		\draw [Green,thick] (30.5,3.5) -- (30.5,5.3);
		\draw [Green,thick] (30.5,5.3) -- (32.3,5.3);
		\draw [Green,thick,-stealth] (32.3,5.3) -- (32.3,1);
		
		\node[scale=2.5, rotate=90] at (0,-7.6) {Cumulative reward};
		\node[scale=2.5] at (8,-13) {Time steps};
		\node[scale=2.5] at (8,-15) {(c)};
		\node[anchor=south west,inner sep=0,scale=0.8] at (1,-12) {\includegraphics[width=\textwidth]{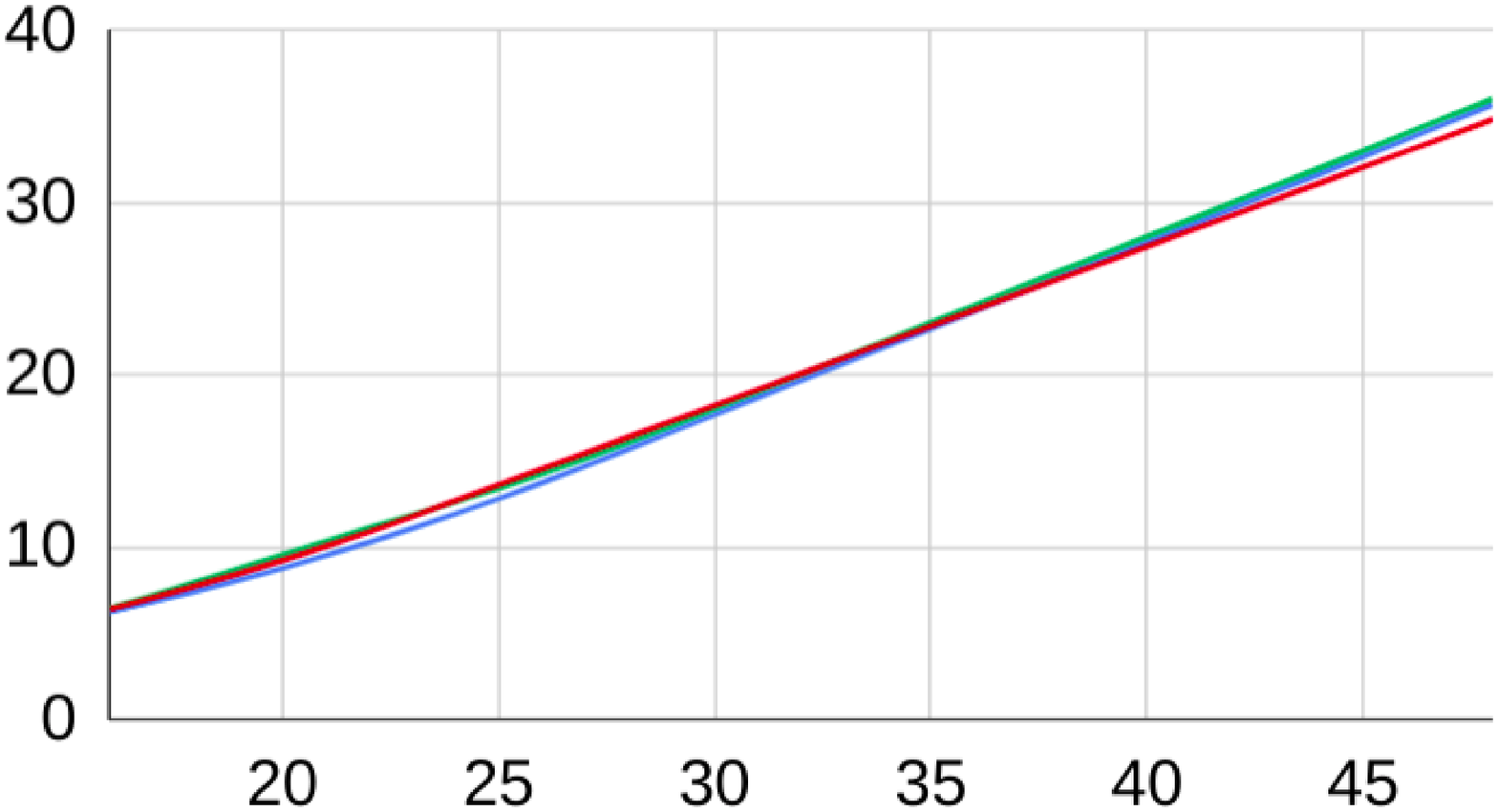}};

		\node[scale=2.5, rotate=90] at (18,-7.6) {Cumulative reward};
		\node[scale=2.5] at (26,-13) {Time steps};
		\node[scale=2.5] at (26,-15) {(d)};
		\node[anchor=south west,inner sep=0,scale=0.8] at (19,-12) {\includegraphics[width=\textwidth]{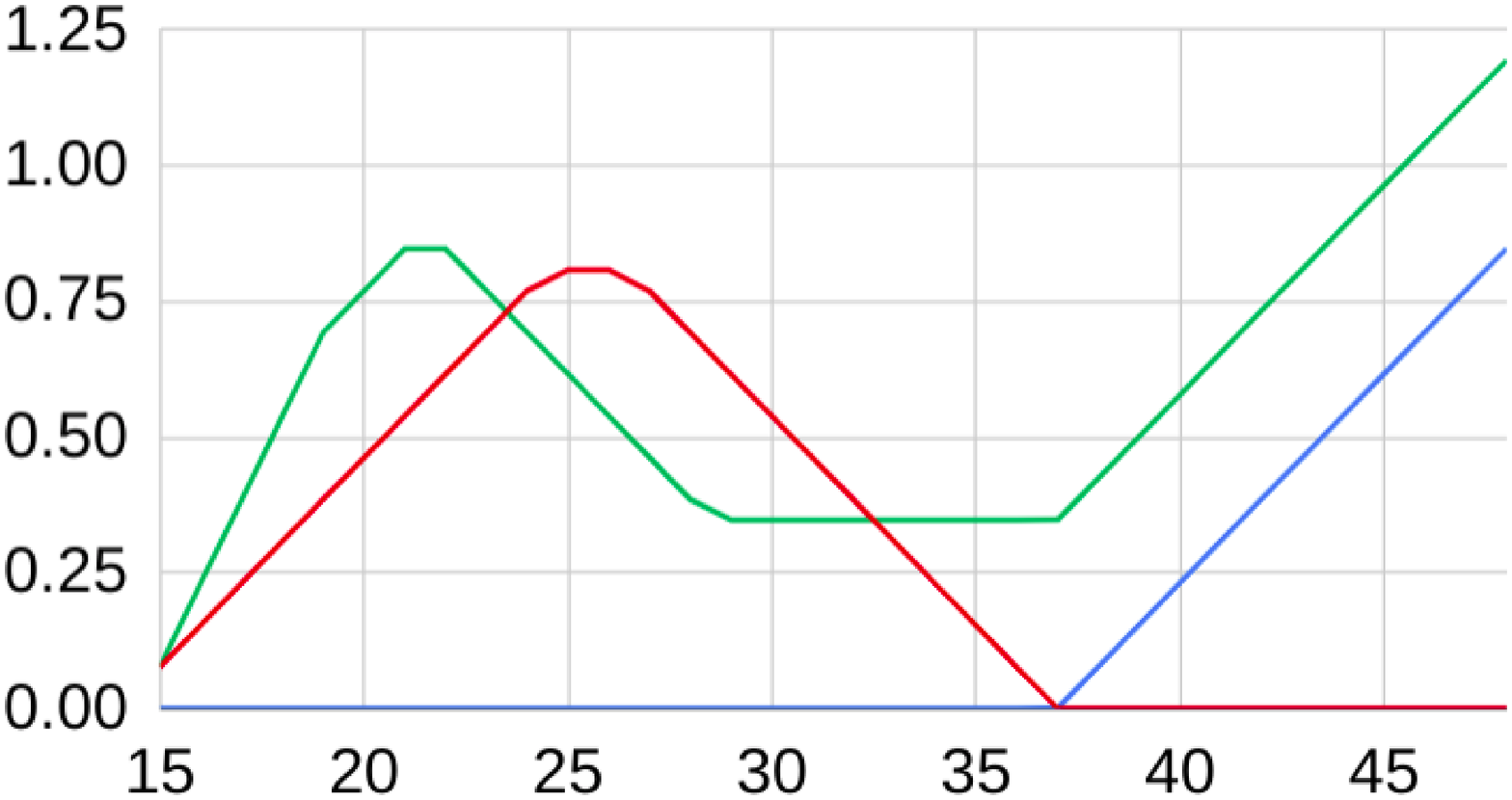}};

    \end{tikzpicture}
	\end{center}
\vspace{-0.25cm}
    \caption{
(a) and (b) illustrate the lakes used to perform the experiments of the present section. The black squares correspond to the external boundary of the lake, the green square corresponds to the frisbee location, the red squares correspond to the agent starting position, the orange squares correspond to local minima of the lake (not all local minima are represented), and the dark blue squares correspond to the holes in which the agent can fall if not careful. Note, these environments contain over 100 states, i.e., one for each cell within the external boundary. Finally, in (b) the green path corresponds to the path taken by the BTAI agent, the red path corresponds to the path selected by the POMCP agent (see the results in the main text), and the blue path corresponds to the shortest path connecting the starting position to the frisbee location. By the ``shortest path", we mean the path that is passing through the smallest number of frozen surfaces without passing through a hole. (c) shows the cumulative reward (CR) received by the agent when following the green, red or blue path. The x-axis corresponds to the number of time steps, i.e., number of action-perception cycles, for which the agent follows the green, red or blue path. We see that all three paths have almost identical values. (d) shows the CR obtained along the green, red and blue paths minus the minimum cumulative reward (MCR) at each time step, where: $\text{MCR} = \min (\text{CR}_{\text{green}},\text{CR}_{\text{red}},\text{CR}_{\text{blue}})$.}
    \label{fig:lakes}
\end{figure}
}

In terms of the reward funtion, the agent receives a penalty of minus one each time it steps on a hole. Otherwise, the agent receives a reward between zero and one. This reward increases linearly as the agent gets closer to the frisbee location, where the distance between the agent and the frisbee is measured using the Manhattan distance as for the maze environment. Note, the reward received by the agent is maximum when the agent stands at the frisbee location, for which it receives a reward of one. Also in theory, this task does not have any terminal states, and the agent will keep taking actions forever. However, in practice, each trial is stopped after a fixed number of action-perception cycles. Figures \ref{fig:lakes}(a) and \ref{fig:lakes}(b) present the lakes in which the upcoming simulations have been ran. For reproducibility, we provide the values of the hyper-parameters used throughout this section in Appendix D.

\subsection{BTAI on the frozen lake environment}

Table \ref{tab:5503} shows the results obtained by the BTAI agent on the lake of Figure \ref{fig:lakes}(a). In short, the BTAI agent required twenty planning iterations before it was able to solve this task. Each simulation takes an average of 7.870 seconds of computational time, which correspond to approximatly $7.870/30 \approx 0.262$ seconds of thinking (i.e., inference, planning and action selection) per action-perception cycle. 

\begin{table}[H]
\centering
\begin{tabular}{ |c|c|c|c| }
 \hline
 Planning iterations & P(global) & P(local) & Time (sec) \\
 \hline
 10 & 0 & 1 & 6.991 $\pm$ 0.459 \\
 \hline
 15 & 0 & 1 & 7.820 $\pm$ 0.577 \\
 \hline
 20 & 1 & 0 & 7.870 $\pm$ 0.707 \\
 \hline
\end{tabular}
\caption{This table presents the probability that the BTAI agent solves the lake of Figure \ref{fig:lakes}(a), and the probability of the agent falling into a local minimum of the EFE. Where by ``falling into a local minimum", we mean that the agent gets stuck into cells of the lake (apart from the exit) for which no adjacent cell represents a frozen surface that has a lower distance to the exit. The last column reports the execution time required for running one simulation and the associated standard deviation.}
\label{tab:5503}
\end{table}

Table \ref{tab:5504} shows the results obtained by the BTAI agent on the lake of Figure \ref{fig:lakes}(b). In short, the BTAI agent requires fifty planning iterations to be able to solve this task. Each simulation takes an average of 19.187 seconds of computational time, which correspond to approximatly $19.187/30 \approx 0.639$ seconds of thinking (i.e., inference, planning and action selection) per action-perception cycle. 

\begin{table}[H]
\centering
\begin{tabular}{ |c|c|c|c| }
 \hline
 Planning iterations & P(global) & P(local) & Time (sec) \\
 \hline
 30 & 0 & 1 & 12.810 $\pm$ 1.071 \\
 \hline
 40 & 0 & 1 & 15.589 $\pm$ 0.766 \\
 \hline
 50 & 1 & 0 & 19.187 $\pm$ 1.317 \\
 \hline
\end{tabular}
\caption{This table presents the probability that the BTAI agent solves the lake of Figure \ref{fig:lakes}(b), and the probability of the agent falling into a local minimum of the EFE. Where by ``falling into a local minimum", we mean that the agent gets stuck into cells of the lake (apart from the exit) for which no adjacent cell represents a frozen surface that has a lower distance to the exit. The last column reports the (average) execution time required for running one simulation, as well as the associated standard deviation.}
\label{tab:5504}
\end{table}

\subsection{POMCP on the frozen lake environment}

In this section, we compare BTAI to the partially observable Monte Carlo planning (POMCP) algorithm introduced by \citet{POMCP}. The code implementing the POMCP algorithm is available at the following URL: \url{https://github.com/ChampiB/POMCP}. Briefly, the POMCP agent performs MCTS \citep{Go,6145622,MuZero} to select an action at each time step, and carries out inference using a particle filter \citep{PARTICLE_FILTER}. Table \ref{tab:5505} shows the results obtained by the POMCP agent on the lake of Figure \ref{fig:lakes}(a). In short, the POMCP agent was able to reach the frisbee 97 \% of the time when using one thousand planning iterations. At which point, each simulation takes an average of 40.444 seconds of computational time, which correspond to approximatly $40.444/30 \approx 1.348$ seconds of thinking (i.e., inference, planning and action selection) per action-perception cycle. This seems to indicate that BTAI is able to solve this first lake four times faster than the POMCP algorithm.

\begin{table}[H]
\centering
\begin{tabular}{ |c|c|c|c| }
 \hline
 Planning iterations & P(global) & P(local) & Time (sec) \\
 \hline
 100 & 0.52 & 0.48 & 3.852 $\pm$ 0.227 \\
 \hline
 500 & 0.89 & 0.11 & 20.550 $\pm$ 3.054 \\
 \hline
 1000 & 0.97 & 0.03 & 40.444 $\pm$ 3.232 \\
 \hline
 2000 & 0.93 & 0.07 & 83.156 $\pm$ 8.844 \\
  \hline
\end{tabular}
\caption{This table presents the probability that the POMCP agent solves the lake of Figure \ref{fig:lakes}(a), and the probability of the agent falling into a local maximum of the reward function. The last column of the above table reports the execution time required for running one simulation and the associated standard deviation. Importantly, this table can be compared with Table \ref{tab:5503} that presents the performance of the BTAI agent on the same lake.}
\label{tab:5505}
\end{table}

On the lake of Figure \ref{fig:lakes}(b), the POMCP agent picks the red path, while the BTAI agent chooses the green path. As shown by Figure \ref{fig:lakes}(c), even if BTAI reaches the goal state while POMCP does not, the cumulative reward obtained by both agents is almost identical. This means that both agents collect a similar amount of reward.

Interestingly, the approach receiving the largest amount of reward depends on the number of time steps in each simulation, i.e., the length of each episode. Figure \ref{fig:lakes}(d) illustrates when BTAI is receiving more rewards than the POMCP algorithm, and when the opposite is true. To sum up, if a simulation is composed of between one and fifteen time step(s), both approaches are equivalent. If a simulation contains between sixteen and twenty-three action-perception cycles, BTAI will accumulate more rewards than the POMCP algorithm. If the simulation has between twenty-four and thirty-two time steps, then the POMCP agent will accumulate more rewards than BTAI. Lastly, if the simulation contains more than twenty-three action-perception cycles, BTAI will accumulate more rewards than the POMCP agent. Thus, in the long run, the POMCP algorithm selects a reasonable but slightly suboptimal path. This might be due to the small difference of cumulated reward obtained along the optimal path and the path taken by the POMCP algorithm. Also, this may be worsened both by the large number of time steps required before to see any difference in accumulated reward between those two paths, and the variance of the MCTS algorithm \citep{veness2011variance}.

Note, the blue path in Figure \ref{fig:lakes}(b) is the shortest path connecting the starting position to the goal state, but is never optimal in terms of cumulative reward. This is because the blue path makes a detour through an area of the lake with low reward, while the green path makes a longer detour but passes through an area with higher rewards. Finally, if the reward received by the agent upon reaching the frisbee (i.e., green square) is increased sufficiently, then the POMCP agent gains incentive to cross the hole separating it from the frisbee, i.e., POMCP will accept a large penalty for an even greater reward.

\section{The dSprites environment}\label{sec:sprites_env}

\begin{figure}[H]
	\begin{center}
	\includegraphics[scale=2]{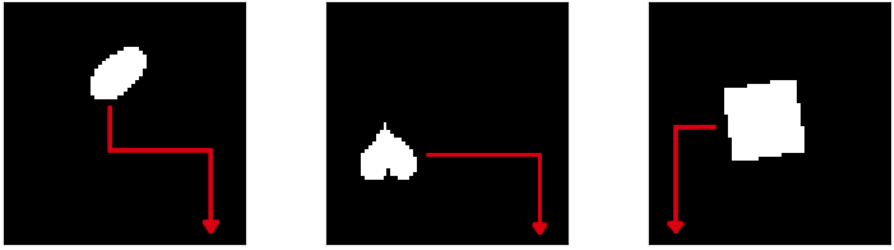}
	\end{center}
  \caption{This figure illustrates the dSprites environment, in which the agent must move all squares towards the bottom-left corner of the image and all ellipses and hearts towards the bottom-right corner of the image. The red arrows show the behaviour expected from the agent.}
   \label{fig:dSprites_env}
\end{figure}

The dSprites environment is based on the dSprites dataset \citep{dsprites17} initially designed for analysing the latent representation learned by variational auto-encoders \citep{VAE}. The dSprites dataset is composed of images of squares, ellipses and hearts. Each image contains one shape (square, ellipse or heart) with its own size, orientation, and $(X,Y)$ position. In the dSprites environment, the agent is able to move those shapes around by performing four actions (i.e., UP, DOWN, LEFT, RIGHT). To make planning tractable, the action selected by the agent is executed eight times in the environment before the beginning of the next action-perception cycle, i.e., the $X$ or $Y$ position is increased or decreased by eight between time step $t$ and $t+1$. The goal of the agent is to move all squares towards the bottom-left corner of the image and all ellipses and hearts towards the bottom-right corner of the image, c.f. Figure \ref{fig:dSprites_env}.

Since, BTAI is a tabular model whose likelihood $P(O_\tau|S_\tau)$ and transition $P(S_{\tau+1}|S_\tau,U_\tau)$ mappings are represented using matrices, the agent does not directly take images as inputs. Instead, the metadata of the dSprites dataset is used to specify the state space. In particular, the agent observes the type of shape (i.e., square, ellipse, or heart), as well as a coarse-grained version of the shape's true position. Importantly, the original images are composed of 32 possible values for both the $X$ and $Y$ positions of the shapes. A coarse-grained  representation with a granularity of two means that the agent is only able to perceive $16 \times 16$ images, and thus, the positions at coordinate $(0,0)$, $(0,1)$, $(1,0)$ and $(1,1)$ are indistinguishable. Figure \ref{fig:down_sampling} illustrates the coarse grained representation with a granularity of eight and the corresponding indices observed by the agent. Note that this modification of the observation space can be seen as a form of state aggregation \citep{STATES_AGGREG}. Finally, as shown in Figure \ref{fig:down_sampling}, the prior preferences of the agent are specified over an imaginary row below the dSprites image. This imaginary row ensures that the agent selects the action ``down" when standing in the ``appropriate corner", i.e., bottom-left corner for squares and bottom-right coner for ellipses and hearts.

{
\colorlet{Green}{green!70!black}
\colorlet{Red}{red!80!black}
\renewcommand\fbox{\fcolorbox{white}{white}}
\begin{figure}[H]
	\begin{center}
	\begin{tikzpicture}[scale=0.4, every node/.style={scale=0.4}]
	\node at (-5.5, 2.5) {\includegraphics[scale=2.5]{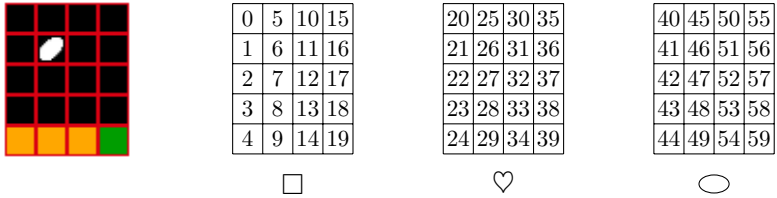}};

	\draw[step=1.0,black,thin] (0,0) grid (4,5);	
	\draw[step=1.0,black,thin] (7,0) grid (11,5);	
	\draw[step=1.0,black,thin] (14,0) grid (18,5);	
	
	\node[scale=2.5] at (2,-1) {$\square$};
	\node[scale=2.5] at (9,-1) {$\heart$};
	\draw (16,-1) ellipse (0.5cm and 0.3cm);

	\node[scale=2] at (0.5,4.5) {0};
	\node[scale=2] at (0.5,3.5) {1};
	\node[scale=2] at (0.5,2.5) {2};
	\node[scale=2] at (0.5,1.5) {3};
	\node[scale=2] at (0.5,0.5) {4};

	\node[scale=2] at (1.5,4.5) {5};
	\node[scale=2] at (1.5,3.5) {6};
	\node[scale=2] at (1.5,2.5) {7};
	\node[scale=2] at (1.5,1.5) {8};
	\node[scale=2] at (1.5,0.5) {9};

	\node[scale=2] at (2.5,4.5) {10};
	\node[scale=2] at (2.5,3.5) {11};
	\node[scale=2] at (2.5,2.5) {12};
	\node[scale=2] at (2.5,1.5) {13};
	\node[scale=2] at (2.5,0.5) {14};

	\node[scale=2] at (3.5,4.5) {15};
	\node[scale=2] at (3.5,3.5) {16};
	\node[scale=2] at (3.5,2.5) {17};
	\node[scale=2] at (3.5,1.5) {18};
	\node[scale=2] at (3.5,0.5) {19};

	\node[scale=2] at (7.5,4.5) {20};
	\node[scale=2] at (7.5,3.5) {21};
	\node[scale=2] at (7.5,2.5) {22};
	\node[scale=2] at (7.5,1.5) {23};
	\node[scale=2] at (7.5,0.5) {24};

	\node[scale=2] at (8.5,4.5) {25};
	\node[scale=2] at (8.5,3.5) {26};
	\node[scale=2] at (8.5,2.5) {27};
	\node[scale=2] at (8.5,1.5) {28};
	\node[scale=2] at (8.5,0.5) {29};

	\node[scale=2] at (9.5,4.5) {30};
	\node[scale=2] at (9.5,3.5) {31};
	\node[scale=2] at (9.5,2.5) {32};
	\node[scale=2] at (9.5,1.5) {33};
	\node[scale=2] at (9.5,0.5) {34};

	\node[scale=2] at (10.5,4.5) {35};
	\node[scale=2] at (10.5,3.5) {36};
	\node[scale=2] at (10.5,2.5) {37};
	\node[scale=2] at (10.5,1.5) {38};
	\node[scale=2] at (10.5,0.5) {39};

	\node[scale=2] at (14.5,4.5) {40};
	\node[scale=2] at (14.5,3.5) {41};
	\node[scale=2] at (14.5,2.5) {42};
	\node[scale=2] at (14.5,1.5) {43};
	\node[scale=2] at (14.5,0.5) {44};
	
	\node[scale=2] at (15.5,4.5) {45};
	\node[scale=2] at (15.5,3.5) {46};
	\node[scale=2] at (15.5,2.5) {47};
	\node[scale=2] at (15.5,1.5) {48};
	\node[scale=2] at (15.5,0.5) {49};
	
	\node[scale=2] at (16.5,4.5) {50};
	\node[scale=2] at (16.5,3.5) {51};
	\node[scale=2] at (16.5,2.5) {52};
	\node[scale=2] at (16.5,1.5) {53};
	\node[scale=2] at (16.5,0.5) {54};
	
	\node[scale=2] at (17.5,4.5) {55};
	\node[scale=2] at (17.5,3.5) {56};
	\node[scale=2] at (17.5,2.5) {57};
	\node[scale=2] at (17.5,1.5) {58};
	\node[scale=2] at (17.5,0.5) {59};
	    \end{tikzpicture}
	\end{center}
\vspace{-0.25cm}
    \caption{
This figure illustrates the observations made by the agent when using a coarse-grained representation with a granularity of eight on the input image. On the left, one can see an image from the dSprites dataset and a grid containing red squares of $8\times8$ pixels. Any positions in those $8\times8$ squares are indistinguishable from the perspective of the agent. Also, the bottom most row is an imaginary row used to specify the prior preferences of the agent, i.e. the green square is the goal state and the orange squares correspond to undesirable states. Finally, the three tables on the right contain the indices observed by the agent for each type of shape at each possible position.}
    \label{fig:down_sampling}
\end{figure}
}

The evaluation of the agent's performance is based on the reward obtained by the agent. Briefly, the agent receives a reward of $-1$, if it never enters the imaginary row or if it does so at the antipode of the appropriate corner. As the agent enters the imaginary row closer and closer to the appropriate corner, its reward increases until reaching a maximum of $1$. The percentage of the task solved (i.e., the evaluation metric) is calculated as follows:
$$P(\text{solved}) = \frac{\text{total rewards} + \text{number of runs}}{2.0 \times \text{number of runs}}.$$
Intuitively, the numerator shifts the rewards so that they are bounded between zero and two, and the denominator renormalises the reward to give a score between zero and one. A score of zero therefore corresponds to an agent always failing to enter the imaginary row or doing so at the antipode of the appropriate corner. In contrast, a score of one corresponds to an agent always entering the imaginary row through the appropriate corner.

\subsection{BTAI on the dSprites environment} \label{ssec:BTAI_dSprites}

In this section, we evaluate BTAI on the dSprites environment. The hyper-parameters used in this section are presented in Appendix D. Briefly, the agent is able to solve 88.5\% of the task when using a granularity of eight, c.f. Table \ref{tab:dSprites_res_8}. To understand why BTAI was not able to solve the task with 100\% accuracy, let us consider the example of an ellipse at position $(24,31)$. With a granularity of eight, the agent perceives that the ellipse is in the bottom-right corner of the image, i.e., in the red square just above the goal state in Figure \ref{fig:down_sampling}. From the agent's perspective, it is thus optimal to pick the action ``down" to reach the goal state. However, in reality, the agent will not reveive the maximum reward because its true $X$ position is $24$ instead of the optimal $X$ position of $31$. 

As shown in Table \ref{tab:dSprites_res_4}, we can improve the agent's perfomance, by using a granularity of four. This allows the agent to differentiate between a larger number of $(X,Y)$ positions, i.e., it reduces the size of the red square in Figure \ref{fig:down_sampling}. With this setting, the agent is able to solve 96.9\% of the task. However, when decreasing the granularity, the number of states goes up, and so does the width and height of the $\bm{A}$ and $\bm{B}$ matrices. As an effect, more memory and computational time is required for the inference and planning process. This highlights a trade-off between the agent's performance and the amount of memory and time required. Indeed, a smaller granularity leads to better performance, but requires more time and memory.

\begin{table}[H]
\centering
\begin{tabular}{ |c|c|c| }
 \hline
 Planning iterations & P(solved) & Time (sec) \\
 \hline
 10 & 0.813	 & 0.859 $\pm$ 0.868 \\
 \hline
 25 & 0.846 & 0.862 $\pm$ 0.958 \\
 \hline
 50 & 0.885 & 1.286 $\pm$ 1.261 \\
 \hline
\end{tabular}
\caption{This table presents the percentage of the dSprites environment solved by the BTAI agent when using a granularity of eight, c.f. Figure \ref{fig:down_sampling}. The last column reports the average execution time required for one simulation and the associated standard deviation.}
\label{tab:dSprites_res_8}
\end{table}

\begin{table}[H]
\centering
\begin{tabular}{ |c|c|c| }
 \hline
 Planning iterations & P(solved) & Time (sec) \\
 \hline
 10 & 0.859 & 3.957 $\pm$ 4.027 \\
 \hline
 25 & 0.933 & 3.711 $\pm$ 4.625 \\
 \hline
 50 & 0.969 & 5.107 $\pm$ 5.337 \\
 \hline
\end{tabular}
\caption{This table presents the percentage of the dSprites environment solved by the BTAI agent when using a granularity of four. In this setting, there are $9 \times 8 \times 3 = 216$ states. The last column reports the average execution time required for one simulation and the associated standard deviation.}
\label{tab:dSprites_res_4}
\end{table}

\subsection{Fountas et al approach on the dSprites environment}

In this section, we experiment with the approach of \citet{DeepAIwithMCMC}. The code used in this section is available on Github at the following URL: \url{https://github.com/ChampiB/deep-active-inference-mc}. First, we trained the agent for around two days on a Nvidia Tesla P100 GPU. After the training process, we ran 100 simulations on the original dSprites environment with both CPU and GPU. Table \ref{tab:dSprites_fountas} reports the percentage of the task solved and the average time required for running a trial. Running the CPU simulations took on average 17.811 seconds per simulation. This is around three times longer than the GPU counterpart, which required an average of 5.467 seconds per simulation. Fountas' agent was able to solve up to 84.1 \% of the task, which is less than the 96.9 \% achieved by the BTAI agent in the previous section.

However, it is important to acknowledge the differences between the present paper and \citet{DeepAIwithMCMC}, as well as the differences between the two environments on which those approaches have been evaluated. First, our approach is not equipped with deep neural networks, and is therefore unable to deal with images as input. Additionally, our agent was not asked to learn the environment's dynamics, instead, our agent was provided with a model of the environment since we are focusing on planning. In contrast, the agent of \citet{DeepAIwithMCMC} was able to successfully learn the environment's dynamics directly from images and then do the planning.

To conclude, our approach was able to solve 96.9 \% of a simplified version of the dSprites environment, and the agent of \citet{DeepAIwithMCMC} was able to solve 84.1 \% of the original dSprites environment. Additionally, our approach was provided with the environment's dynamics, while the agent of \citet{DeepAIwithMCMC} had to learn it, which took around two days on a Nvidia Tesla P100 GPU. Another, important trade-off is between interpretability and scalability. Indeed, the tabular representation of the likelihood and transition mappings makes the BTAI agent very intuitive and easy to understand. However, this tabular representation is also the main bottleneck blocking BTAI from solving image based environments. Similarly, the deep neural networks used by \citet{DeepAIwithMCMC} make their approach highly scalable, but also reduce the interpretability of the approach.

\begin{table}[H]
\centering
\begin{tabular}{ |c|c|c| }
 \hline
 Computation type & P(solved) & Time (sec) \\
 \hline
 CPU & 0.798 & 17.811 $\pm$ 19.143\\
 GPU & 0.841 & 5.467 $\pm$ 5.706\\
 \hline
\end{tabular}
\caption{This table presents the percentage of the original dSprites environment solved by the approach of \citet{DeepAIwithMCMC}. The last column reports the average execution time required for one simulation and the associated standard deviation. Importantly, this table can be compared with Table \ref{tab:dSprites_res_4} that presents the performance of the BTAI agent on a simplified version of the dSprites environment.}
\label{tab:dSprites_fountas}
\end{table}

\section{Conclusion and future works} \label{sec:conclusion}

In this paper, we provided an empirical study of branching time active inference (BTAI), where the name takes inspiration from branching-time theories of concurrent and distributed systems in computer science \citep{concurrency_glabbeek,concurrency_glabbeek_2,concurrency_howard}, and planning was cast as (temporal) structure learning. Simply put, the generative model is dynamically expanded and each expansion leads to the exploration of new policy fragments. The expansions are guided by the expected free energy, which provides a trade off between exploration and exploitation. Importantly, this approach is composed of not two, but three major distributions. The first is the prior distribution (or generative model) that encodes the agent's beliefs before performing any observations. The second is the posterior (or variational) distribution encoding the updated beliefs of the agent after performing some observations. And the third is a target distribution over future states and observations that encodes the prior preferences of the agent, i.e., a generalization of the $\bm{C}$ matrix in the standard formulation of active inference proposed by \citet{FRISTON2016862}. An important advantage of this generalization is that it allows the specification of prior preferences over both future observations and future states at the same time.

We compared BTAI and standard active inference theoretically by studying its space and time complexity class. This study highlights that our method should perform better than the standard model used in active inference when the task can be solved by expanding the tree only a small number of times with respect to an exhaustive search. Second, we compared BTAI to active inference empirically within the deep reward environment. Those simulations suggest that BTAI is able to solve problems for which a standard active inference agent would run out of memory. Interestingly, active inference offers an Occam's window \citep{AI_TUTO} for policy pruning, i.e., a policy is prunned if its posterior probability is very low w.r.t. the current best policy. This approach provides a way to reduce the amount of space used by active inference, since the policies with low probability and their associated beliefs over states can be discarded. However, a direct application of Occam's window will not solve the exponential time complexity class because the posterior probability of all policies still needs to be evaluated. It seems that a new AcI-based algorithm would be required to use the potential of Occam's window. As elaborated upon in Section \ref{ssec:complexity_class}, one might argue that there is a trade-off between banching-time active inference, which provides considerably more efficient planning to attain current goals, and classic active inference which provides a more exhaustive assessment of paths not taken. This might enable active inference to more exhaustively reflect counter-factuals and reasoning based upon them.

Also, BTAI was studied (experimentally) in the context of a maze solving task and we showed that when the heuristic used to create the prior preferences is not perfect, the agent becomes vulnerable to local minima. In other words, the agent might be attracted by a part of the maze that has low cost but does not allow it to solve the task. Then, we demonstrated empirically that improving the prior preferences of the agent by specifying a good prior over future hidden states and deepening the tree search, helped to mitigate this issue.

Moreover, BTAI was compared to the POMCP algorithm \citep{POMCP} on the frozen lake environment. This comparison was based upon two lakes each having their own topology. In terms of performance, both approaches successfully solved the simplest lake. On the hardest lake, BTAI and the POMCP algorithm received a similar amount of reward. Also, we described when BTAI receives more rewards than the POMCP agent, and when the opposite is true.

Additionally, BTAI was compared to the approach of \citet{DeepAIwithMCMC} on the dSprites dataset. The experiments show that our approach was able to solve 96.9 \% of a simplified version of the dSprites environment, and the agent of \citet{DeepAIwithMCMC} was able to solve 84.1 \% of the original dSprites environment. However, our approach was provided with the environment's dynamics, while the agent of \citet{DeepAIwithMCMC} had to learn it, which took around two days on a Nvidia Tesla P100 GPU. Another, important trade-off is between interpretability and scalability. Indeed, the tabular representation of the likelihood and transition mappings makes the BTAI agent very intuitive and easy to understand. Unfortunatly, this tabular representation is also the main bottleneck blocking BTAI from solving image based environments. Similarly, the deep neural networks used by \citet{DeepAIwithMCMC} make their approach highly scalable, but reduce the interpretability of this approach.

The present paper could lead to a large number of future research directions. One could for example add the ability of the agent to learn the transition matrices $\bm{B}$ as well as the likelihood matrix $\bm{A}$ and the vector of initial states $\bm{D}$. This can be done in at least two ways. The first is to add Dirichlet priors over those matrices/vectors and the second would be to use neural networks as function approximators. The second option will lead to a deep active inference agent \citep{PixelBasedAI,DeepAI} equiped with tree search that could be directly compared to the method of \citet{DeepAIwithMCMC}. Including deep neural networks in the framework will also open the door to direct comparison with the deep reinforcement learning literature \citep{DBLP:journals/corr/abs-1801-01290,DeepRL,DDQN,lample2016playing,Go}. Those comparisons will enable the study of the impact of the epistemic terms when the agent is composed of deep neural networks.

Another, important direction of research would be to learn the prior preferences of the agent \citep{sajid2021exploration}. Those preferences are encoded by the vector $\bm{C}$, and could be learned by incorporating a Dirichlet prior over $\bm{C}$. Also, the incorporation of this Dirichlet prior leads to an augmented EFE that could be compared with the standard formulation of the EFE.

Moreover, while the present paper is based on standard active inference that advocates that actions maximize both reward and information gain, it would be interresting to design a version of BTAI based on meta-control \citep{Markovic2021}. Meta-control is a hierarchical model where higher-level hidden states constrain decision making at lower levels. Interestingly, \citet{Markovic2021} argue that it may be beneficial for the agent to switch on and off its exploration tendency based on the current context.

Another direction of research will be to set up behavioural experiments to try to determine which kind of planning is used by the brain. This could simply be done by looking at the time required by a human to solve various mazes and compare it with both the classic model and the tree search alternative. Finally, one could also set up a hierarchical model of action and compare it to the tree search algorithm presented here. One could also evaluate the plausibility of a hierarchical model of action by running behavioural experiments on humans.

Finally, a completely different direction will be to focus on the integration of memory. At the moment, when a new action is performed in the environment and a new observation is recieved from it, all the branches in the tree are prunned and a new temporal slice (i.e. a new state, action and observation triple) is added to the POMDP. In other words, the integration function simply records the past. This exact recording of the past is very unlikely to really happen in the brain. Therefore, one might simply ask what to do with this currently ever growing record of the past. This would certainly lead to the notion of an active inference agent equipped with episodic memory \citep{BOTVINICK2019408}.


\acks{We would like to thank the reviewers for their valuable feedback, which greatly improved the quality of the present paper.}

\appendix

\section*{Appendix A: The theoretical approach of this paper.}

This appendix describes the generative model, the variational distribution and the update equations used throughout this paper. For full details of vocabulary and notation the reader is referred to \citet{AITS_THEORY}.

The generative model can be understood as a fixed part modelling the past and present, and an expandable part modelling the future. The past and present is represented as a sequence of hidden states, where the transition between any two consecutive states depends on the action performed and is modelled using the 3-tensor $\bm{B}$. The generation of an observation is modelled by the matrix $\bm{A}$, and the prior over the initial hidden state as well as the prior over the various actions are modelled using vectors, i.e., $\bm{D}$ and $\bm{\Theta}_\tau$, respectively.

Concerning the second part of the model (i.e., the one modelling the future), the transition between consecutive states in the future is defined using the 2-sub-tensor $\bm{B}(\bigcdot,\bigcdot, I_{\text{last}})$, which is the matrix corresponding to the last action performed to reach the node $S_I$. The generation of future observations from future hidden states is identical to the one used for the past and present.

For the sake of simplicity, we assume that the tensors $\bm{A}$, $\bm{B}$, $\bm{D}$ and $\bm{\Theta}_\tau$ are given to the agent, which means that the agent knows the dynamics of the environment (c.f., Table \ref{tab:active_inf_notation} for additional information about those tensors). Practically, this means that the generative model does not have Dirichlet priors over those tensors. Furthermore, we follow \citet{Parr304782}, by viewing future observations as latent random variables. The formal definition of the generative model, which encodes our prior knowledge of the task, is given by:
\begin{align*}
P(O_{0:t},S_{0:t},U_{0:t-1},O_{\mathbb{I}},S_{\mathbb{I}}) = &P(S_0) \prod_{\tau = 0}^{t - 1} P(U_\tau) \prod_{\tau = 0}^t P(O_\tau|S_\tau)  \prod_{\tau = 1}^t P(S_\tau|S_{\tau - 1}, U_{\tau - 1})\\
&\prod_{I \in \mathbb{I}} P(O_I|S_I)P(S_I|S_{I \setminus \text{last}})
\end{align*}
where $\mathbb{I}$ is the set of all non-empty multi-indexes already expanded, and $S_{I \setminus \text{last}}$ is the parent of $S_I$. Additionally, we need to define the individual factors:
\begin{align*}
&P(S_0) = \text{Cat}(\bm{D})& &P(U_\tau) = \text{Cat}(\bm{\Theta}_\tau) \\
&P(O_\tau|S_\tau) = \text{Cat}(\bm{A})& &P(O_I|S_I) = \text{Cat}(\bm{A}) \\
&P(S_\tau|S_{\tau - 1}, U_{\tau - 1}) = \text{Cat}(\bm{B})& &P(S_I|S_{I \setminus \text{last}}) = \text{Cat}(\bm{B}[I_{\text{last}}]).
\end{align*}
where $I_{last}$ is the last index of the multi-index $I$, i.e., the last action that led to $S_I$, and $\bm{B}[I_{\text{last}}] = \bm{B}(\bigcdot, \bigcdot, I_{\text{last}})$ is the matrix corresponding to $I_{last}$. We now turn to the definition of the variational posterior. Under the mean-field approximation:
\begin{align*}
Q(S_{0:t},U_{0:t-1},O_{\mathbb{I}},S_{\mathbb{I}}) = \prod_{\tau = 0}^{t - 1} Q(U_\tau) \prod_{\tau = 0}^t Q(S_\tau) \prod_{I \in \mathbb{I}} Q(O_I)Q(S_I)
\end{align*}
where the individual factors are defined as:
\begin{align*}
&Q(S_\tau) = \text{Cat}(\bm{\hat{D}}_\tau)& &Q(U_\tau) = \text{Cat}(\bm{\hat{\Theta}}_\tau) \\
&Q(O_I) = \text{Cat}(\bm{\hat{E}}_I)& &Q(S_I) = \text{Cat}(\bm{\hat{D}}_I)
\end{align*}
Lastly, we follow \citet{millidge2020expected} in assuming that the agent aims to minimise the KL divergence between the variational posterior and a desired (target) distribution. Therefore, our framework allows for the specification of prior preferences over both future hidden states and future observations:
\begin{align*}
V(O_{\mathbb{I}},S_{\mathbb{I}}) = \prod_{I \in \mathbb{I}} V(O_I)V(S_I)
\end{align*}
where the individual factors are defined as:
\begin{align*}
V(O_I) = \text{Cat}(\bm{C}_O),& &V(S_I) = \text{Cat}(\bm{C}_S).
\end{align*}
Importantly, $\bm{C}_O$ and $\bm{C}_S$ play the role of the vector $\bm{C}$ in the active inference model \citep{FRISTON2016862}, i.e., they specify which observations and hidden states are rewarding. To sum up, this framework is defined using three distributions: the prior defines the agent's beliefs before performing any observation, the posterior is an updated version of the prior that takes into account the observation made by the agent, and the target (desired) distribution encodes the agent's prior preferences in terms of future observations and hidden states.

\begin{table}[H]
\centerline{
\begin{tabular}{cl}
\hline 
\\[-0.25cm]
Notation & Meaning \\
\\[-0.25cm]
\hline
\hline
\\[-0.4cm]
$T$, $t$ & The time horizon and the current time step\\
\\[-0.4cm]
\hline
\\[-0.4cm]
$O_{i:j}$, $S_{i:j}$, $U_{i:j}$ & The set of observations, states and actions between time step $i$ and $j$ (inclusive)\\
\\[-0.4cm]
\hline
\\[-0.4cm]
$\bm{A}$ & The matrix defining the mapping from states to observations\\
\\[-0.4cm]
\hline
\\[-0.4cm]
\multirow{2}{*}{$\bm{B}$/$\bm{\hat{D}}_\tau$}
& The $3$-tensor defining the mappings (a priori) between any two \\
& consecutive hidden states and the parameters of the posterior over $S_\tau$\\
\\[-0.4cm]
\hline
\\[-0.4cm]
$\bm{D}$/$\bm{\hat{D}}_0$ & The parameters of the prior/posterior over the initial hidden states\\
\\[-0.4cm]
\hline
\\[-0.4cm]
$\bm{\hat{D}}_I$/$\bm{\hat{E}}_I$ & The parameters of the posterior over future states/observations\\
\\[-0.4cm]
\hline
\\[-0.4cm]
$\bm{C}_S$/$\bm{C}_O$ & The parameters of the prior preferences over future states/observations\\
\\[-0.4cm]
\hline
\\[-0.4cm]
$\bm{\Theta}_\tau$/$\bm{\hat{\Theta}}_\tau$ & The parameters of the prior/posterior over actions at time step $\tau$\\
\\[-0.4cm]
\hline
\\[-0.4cm]
$\sigma(\bigcdot)$ & The softmax function\\
\\[-0.4cm]
\hline
\\[-0.4cm]
$\text{Cat}(\bigcdot)$ and $\text{Dir}(\bigcdot)$ & Categorical and Dirichlet distributions\\
\\[-0.4cm]
\hline
\\[-0.4cm]
\end{tabular}
}
\caption{Branching time active inference notation}\label{tab:active_inf_notation}
\end{table}

Finally, the update equations used in this paper rely on variational message passing as presented in \citep{AI_VMP,VMP_TUTO} and are given by:
\begin{align*}
Q^*(S_\tau) = \sigma\Big(&[\tau = 0] \ln \bm{D} \,\,+\,\, [\tau \neq 0] \ln \bm{B} \odot [ \bm{\hat{D}}_{\tau - 1}, \bm{\hat{\Theta}}_{\tau - 1}] {\color{white}\sum^{t}}\\
+ &\ln \bm{A} \odot \bm{o}_\tau {\color{white}\sum_{\tau}^{t}}\\
+ &[\tau = t] \sum_{J \in \text{ch}_t} \ln \bm{B}[J_{last}] \odot \bm{\hat{D}}_{J} \,\,+\,\, [\tau \neq t] \ln \bm{B} \odot [ \bm{\hat{D}}_{\tau + 1}, \bm{\hat{\Theta}}_{\tau}]\Big)\quad \quad \quad \quad \quad \quad \,\,
\end{align*}
\vspace{-1.5cm}
\begin{align*}
Q^*(U_\tau) &= \sigma \big( \ln \bm{\Theta}_\tau + \ln \bm{B} \odot [ \bm{\hat{D}}_{\tau}, \bm{\hat{D}}_{\tau + 1}] \big) {\color{white}\sum_{\tau}^{t}}\\
Q^*(O_I) &= \sigma \big( \ln \bm{A} \odot \bm{\hat{D}}_I \big){\color{white}\sum_{\tau}^{t}}\\
Q^*(S_I) &= \sigma \big( \ln \bm{A} \odot \bm{\hat{E}}_I + \ln \bm{B}[I_{last}] \odot \bm{\hat{D}}_{I \setminus \text{last}} + \sum_{S_K \in \text{ch}_I} \ln \bm{B}[K_{last}] \odot \bm{\hat{D}}_{K} \big) {\color{white}\sum_{\tau}^{t}}
\end{align*}
where $\bm{o}_\tau$ is the observation made at time step $\tau$, $I_{last}$ is the last action of the sequence $I$, $\text{ch}_t$ is the set of multi-indices corresponding to the children of the root node, and $\text{ch}_I$ is the set of multi-indices corresponding to the children of $S_I$. For additional information about $\odot$, the reader is referred to Appendix C.

\section*{Appendix B: Derivation of $g^{pcost}_J$.}

In this appendix, we provide a derivation of $g^{pcost}_J$ from the Free Energy of the Expected Future (FEEF) introduced by \citet{millidge2020expected}:
\begin{equation*}
g^{feef}_I = \kl{Q(O_I, S_I)}{V(O_I, S_I)},
\end{equation*}
by assuming the following factorizations for the variational posterior:
\begin{align*}
Q(O_I,S_I) = Q(O_I)Q(S_I),
\end{align*}
and target distribution:
\begin{align*}
V(O_I,S_I) = V(O_I)V(S_I).
\end{align*}
Starting from $g^{feef}_I$, we use the definition of the KL divergence, the linearity of the expectation, the log property $\ln(ab) = \ln(a) + \ln(b)$, and the two assumptions described above to get:
\begin{align*}
g^{feef}_I &= \kl{Q(O_I, S_I)}{V(O_I, S_I)} &\\
&= \kl{Q(O_I)Q(S_I)}{V(O_I)V(S_I)} &(\text{factorization assumptions})\\
&= \mathbb{E}_{Q(O_I)Q(S_I)}\big[\ln Q(O_I)Q(S_I) - \ln V(O_I)V(S_I)\big] &(\text{KL divergence definition})\\
&= \mathbb{E}_{Q(O_I)Q(S_I)}\big[\ln Q(O_I) - \ln V(O_I) + \ln Q(S_I) - \ln V(S_I)\big] &(\text{log property})\\
&= \mathbb{E}_{Q(O_I)}\big[\ln Q(O_I) - \ln V(O_I)\big] + \mathbb{E}_{Q(S_I)}\big[\ln Q(S_I) - \ln V(S_I)\big] &(\text{linearity of expectation})\\
&= \kl{Q(O_I)}{V(O_I)} + \kl{Q(S_I)}{V(S_I)} &(\text{KL divergence definition})\\
&= g^{pcost}_J.&
\end{align*}

\section*{Appendix C: Generalized inner product}

\paragraph{Generalized inner products:}

\begin{figure}[H]
	\begin{center}
	\includegraphics[scale=0.8]{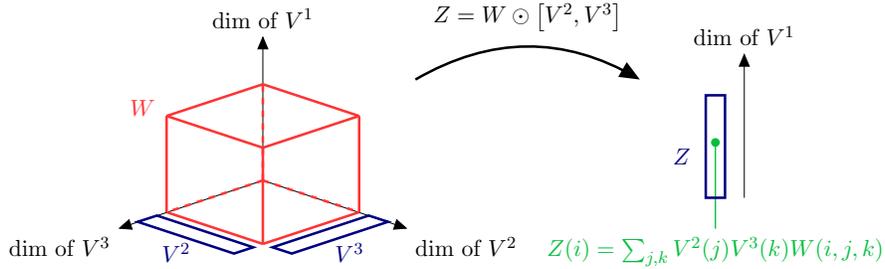}
	\end{center}
  \caption{This figure illustrates the generalized inner product $Z = W \odot \big[ V^2, V^3 \big]$, where $W$ is a cube of values illustrated in red with typical element $W(i,j,k)$. Also, the vectors $Z$ and $V^i \,\, \forall i \in \{2,3\}$ are drawn in blue along the dimension of the cube they correspond to.}
   \label{fig:inner-product}
\end{figure}

Given an $N$ dimensional tensor $W$ and $M = N - 1$ vectors $V^i$, the generalized inner product returns a vector $Z$ obtained by performing a weighted average (with weighting coming from the vectors) over all but one dimension. In other words:
\begin{align*}
Z = W \odot \Big[ V^1, ..., V^{M}\Big] \Leftrightarrow Z(x_j) = \sum_{\substack{x_1 \in \{1, ..., |V^1|\}\\
{\color{white}\}}...{\color{white}\}}\\
x_M \in \{1, ..., |V^M|\} }} V^1_{x_1} \times ... \times W(x_1, ..., & x_j, ..., x_{M}) \times ... \times V^{M}_{x_{M}}\\
& \forall x_j \in \{1, ..., |Z|\},
\end{align*}
where $|Z|$ denotes the number of elements in $Z$, and the large summand is over all $x_r$ for $r \in \{1, ..., M\} \setminus \{j\}$, i.e., excluding $j$. Also, note that if $|W|_{V^i} \,\, \forall i \in \{1, ..., M\}$ is the number of elements in the dimension corresponding to $V^i$, then for $W \odot \big[ V^1, ..., V^{M}\big]$ to be properly defined, we must have $|W|_{V^i} = |V^i| \,\, \forall i \in \{1, ..., M\}$ where $|V^i|$ is the number of elements in $V^i$. Figure \ref{fig:inner-product} illustrates the generalized inner product for $N = 3$.

\paragraph{Naming of the dimensions:}

Importantly, we should imagine that each side of $W$ has a name, e.g., if $W$ is a 3x2 matrix, then the $i$-th dimension of $W$ could be named: ``the dimension of $V_i$". This enables us to write: $Z^1 = W \odot V^1$ and $Z^2 = W \odot V^2$, where $Z^1$ is a 1x2 matrix (i.e., a vector with two elements) and $Z^2$ is a 3x1 matrix (i.e., a vector with three elements). The operator $\odot$ knows (thanks to the dimension name) that $W \odot V^1$ takes the weighted average w.r.t ``the dimension of $V_1$", while $W \odot V^2$ must take the weighted average over ``the dimension of $V_2$".

In the context of active inference, the matrix $\bm{A}$ has two dimensions that we could call ``the observation dimension" (i.e., row-wise) and ``the state dimension" (i.e., column-wise). Trivially, $\bm{A} \odot \bm{o}_\tau$ will then correspond to the average of $\bm{A}$ along the observation dimension and $\bm{A} \odot \bm{\hat{D}}_\tau$ will correspond to the average of $\bm{A}$ along the state dimension. 

\section*{Appendix D: Hyper-parameters used during the simulations}

\paragraph{Lists of hyper-parameters:}

Table \ref{tab:hyper_param_descr} describes the role of the hyper-parameters of the BTAI simulation.

\begin{table}[H]
\centering
\begin{tabular}{ |c|p{12cm}|  }
 \hline
 Name & Description\\
 \hline
 \hline
 \url{NB_SIMULATIONS} & The number of simulations run during the experiment.\\
 \hline
 \url{NB_ACTION_PERCEPTION_CYCLES} & The maximum number of actions in each simulation, after which the simulation is terminated.\\
 \hline
 \url{NB_PLANNING_STEPS} & The number of planning iterations performed by the agent.\\
 \hline
 \url{EXPLORATION_CONSTANT} & The exploration constant of the UCT criterion.\\
 \hline
 \url{PRECISION_PRIOR_PREFERENCES} & The precision of the prior preferences, i.e., $\gamma$ in $\bm{C}_O = \sigma(\gamma \bm{v})$, where $\bm{v}$ is a vector quantifying the preferences of the agent.\\
 \hline
 \url{PRECISION_ACTION_SELECTION} & The precision of the distribution used for action selection, i.e., $\omega$ in $\sigma(-\omega \frac{g}{N})$ where $g$ is a vector whose elements correspond to the cost of the root's children (i.e. the children of $S_t$) and $N$ is a vector whose elements correspond to the number of visits of the root's children.\\
 \hline
 \url{EVALUATION_TYPE} & The type of cost used to evaluate the node during the tree search, i.e., $G^{\text{classic}}_I$ reported as EFE or $G^{\text{pcost}}_I$ reported as DOUBLE\_KL.\\
 \hline
\end{tabular}
\caption{This table describes the hyper-parameters of the BTAI simulation.}
\label{tab:hyper_param_descr}
\end{table}

Table \ref{tab:POMCP_hyper_param_descr} describes the role of the hyper-parameters of the POMCP simulation.

\begin{table}[H]
\centering
\begin{tabular}{ |c|p{12cm}|  }
 \hline
 Name & Description\\
 \hline
 \hline
 \url{NB_SIMULATIONS} & The number of simulations run during the experiment.\\
 \hline
 \url{NB_ACTION_PERCEPTION_CYCLES} & The maximum number of actions in each simulation, after which the simulation is terminated.\\
 \hline
 \url{TIMEOUT} & The number of planning iterations performed by the agent.\\
 \hline
 \url{EXP_CONST} & The exploration constant of the UCT criterion.\\
 \hline
 \url{GAMMA} & The value of the discount factor.\\
 \hline
 \url{NO_PARTICLES} & The number of particles in the filter.\\
 \hline
\end{tabular}
\caption{This table describes the hyper-parameters of the POMCP simulation.}
\label{tab:POMCP_hyper_param_descr}
\end{table}

\paragraph{Hyper-parameters used by BTAI in section \ref{ssec:BTAI_vs_AI_simul}:}

Table \ref{tab:values_hp_ssec_BTAI_vs_AI_simul} provides the value of each hyper-parameter used by BTAI in section \ref{ssec:BTAI_vs_AI_simul}.

\begin{table}[H]
\centering
\begin{tabular}{ |c|c|  }
 \hline
 Name & Value\\
 \hline
 \hline
 \url{NB_SIMULATIONS} & 100\\
 \hline
 \url{NB_ACTION_PERCEPTION_CYCLES} & 20\\
 \hline
 \url{NB_PLANNING_STEPS} & 10 or 15 or 20\\
 \hline
 \url{EXPLORATION_CONSTANT} & 2.4\\
 \hline
 \url{PRECISION_PRIOR_PREFERENCES} & 3\\
 \hline
 \url{PRECISION_ACTION_SELECTION} & 100\\
 \hline
 \url{EVALUATION_TYPE} & EFE\\
 \hline
\end{tabular}
\caption{This table presents the value of each hyper-parameter used by BTAI in section \ref{ssec:BTAI_vs_AI_simul}.}
\label{tab:values_hp_ssec_BTAI_vs_AI_simul}
\end{table}

\paragraph{Hyper-parameters used by BTAI in section \ref{sec:pp_and_lm}:}

Table \ref{tab:values_hp_ssec_pp_and_lm} provides the value of each hyper-parameter used by BTAI in section \ref{sec:pp_and_lm}.

\begin{table}[H]
\centering
\begin{tabular}{ |c|c|  }
 \hline
 Name & Value\\
 \hline
 \hline
 \url{NB_SIMULATIONS} & 100\\
 \hline
 \url{NB_ACTION_PERCEPTION_CYCLES} & 20\\
 \hline
 \url{NB_PLANNING_STEPS} & 10 or 15 or 20\\
 \hline
 \url{EXPLORATION_CONSTANT} & 2.4\\
 \hline
 \url{PRECISION_PRIOR_PREFERENCES} & 2\\
 \hline
 \url{PRECISION_ACTION_SELECTION} & 100\\
 \hline
 \url{EVALUATION_TYPE} & EFE\\
 \hline
\end{tabular}
\caption{This table presents the value of each hyper-parameter used by BTAI in section \ref{sec:pp_and_lm}.}
\label{tab:values_hp_ssec_pp_and_lm}
\end{table}

\paragraph{Hyper-parameters used by BTAI in section \ref{sec:pp_and_lm_mp}:}

Table \ref{tab:values_hp_pp_and_lm_mp} provides the value of each hyper-parameter used by BTAI in section \ref{sec:pp_and_lm_mp}.

\begin{table}[H]
\centering
\begin{tabular}{ |c|c|  }
 \hline
 Name & Value\\
 \hline
 \hline
 \url{NB_SIMULATIONS} & 100\\
 \hline
 \url{NB_ACTION_PERCEPTION_CYCLES} & 20\\
 \hline
 \url{NB_PLANNING_STEPS} & 10 or 15 or 20\\
 \hline
 \url{EXPLORATION_CONSTANT} & 2.4\\
 \hline
 \url{PRECISION_PRIOR_PREFERENCES} & 2\\
 \hline
 \url{PRECISION_ACTION_SELECTION} & 100\\
 \hline
 \url{EVALUATION_TYPE} & DOUBLE\_KL\\
 \hline
\end{tabular}
\caption{This table presents the value of each hyper-parameter used by BTAI in section \ref{sec:pp_and_lm_mp}.}
\label{tab:values_hp_pp_and_lm_mp}
\end{table}

\paragraph{Hyper-parameters used by BTAI in section \ref{sec:results_costs}:}

Table \ref{tab:values_hp_results_costs} provides the value of each hyper-parameter used by BTAI in section \ref{sec:results_costs}.

\begin{table}[H]
\centering
\begin{tabular}{ |c|c|  }
 \hline
 Name & Value\\
 \hline
 \hline
 \url{NB_SIMULATIONS} & 100\\
 \hline
 \url{NB_ACTION_PERCEPTION_CYCLES} & 20\\
 \hline
 \url{NB_PLANNING_STEPS} & 10 or 15 or 20\\
 \hline
 \url{EXPLORATION_CONSTANT} & 2.4\\
 \hline
 \url{PRECISION_PRIOR_PREFERENCES} & 2\\
 \hline
 \url{PRECISION_ACTION_SELECTION} & 100\\
 \hline
 \url{EVALUATION_TYPE} & EFE or DOUBLE\_KL\\
 \hline
\end{tabular}
\caption{This table presents the value of each hyper-parameter used by BTAI in section \ref{sec:results_costs}.}
\label{tab:values_hp_results_costs}
\end{table}

\paragraph{Hyper-parameters used by BTAI in section \ref{sec:frozen_lake_env}:}

Table \ref{tab:values_hp_frozen_lake_env} provides the value of each hyper-parameter used by BTAI in section \ref{sec:frozen_lake_env}.

\begin{table}[H]
\centering
\begin{tabular}{ |c|c|  }
 \hline
 Name & Value\\
 \hline
 \hline
 \url{NB_SIMULATIONS} & 100\\
 \hline
 \url{NB_ACTION_PERCEPTION_CYCLES} & 30\\
 \hline
 \url{NB_PLANNING_STEPS} & 10, 15, 20, 30, 40 or 50\\
 \hline
 \url{EXPLORATION_CONSTANT} & 2.4\\
 \hline
 \url{PRECISION_PRIOR_PREFERENCES} & 2\\
 \hline
 \url{PRECISION_ACTION_SELECTION} & 100\\
 \hline
 \url{EVALUATION_TYPE} & EFE\\
 \hline
\end{tabular}
\caption{This table presents the value of each hyper-parameter used by BTAI in section \ref{sec:frozen_lake_env}. Note, the number of action-perception cycles has been increased from 20 to 30, because the agent cannot possibly solve the task with 20 actions (the lake is too large).}
\label{tab:values_hp_frozen_lake_env}
\end{table}

\paragraph{Hyper-parameters used by the POMCP algorithm in section \ref{sec:frozen_lake_env}:}

Table \ref{tab:values_hp_frozen_lake_env_POMCP} provides the value of each hyper-parameter used by the POMCP algorithm in section \ref{sec:frozen_lake_env}.

\begin{table}[H]
\centering
\begin{tabular}{ |c|c|  }
 \hline
 Name & Value\\
 \hline
 \hline
 \url{NB_SIMULATIONS} & 100\\
 \hline
 \url{NB_ACTION_PERCEPTION_CYCLES} & 30\\
 \hline
 \url{TIMEOUT} & 100, 500, 1000 or 2000\\
 \hline
 \url{EXP_CONST} & 3\\
 \hline
 \url{GAMMA} & 0.9\\
 \hline
 \url{NO_PARTICLES} & 100\\
 \hline
\end{tabular}
\caption{This table presents the value of each hyper-parameter used by the POMCP algorithm in section \ref{sec:frozen_lake_env}.}
\label{tab:values_hp_frozen_lake_env_POMCP}
\end{table}

\paragraph{Hyper-parameters used by BTAI in section \ref{ssec:BTAI_dSprites}:}

Table \ref{tab:values_hp_BTAI_dSprites} provides the value of each hyper-parameter used by BTAI in section \ref{ssec:BTAI_dSprites}. Also, note that the granularity of the coarse-grained representation was set to four or eight.

\begin{table}[H]
\centering
\begin{tabular}{ |c|c|  }
 \hline
 Name & Value\\
 \hline
 \hline
 \url{NB_SIMULATIONS} & 100\\
 \hline
 \url{NB_ACTION_PERCEPTION_CYCLES} & 30\\
 \hline
 \url{NB_PLANNING_STEPS} & 10, 25 or 50\\
 \hline
 \url{EXPLORATION_CONSTANT} & 2.4\\
 \hline
 \url{PRECISION_PRIOR_PREFERENCES} & 2\\
 \hline
 \url{PRECISION_ACTION_SELECTION} & 100\\
 \hline
 \url{EVALUATION_TYPE} & EFE\\
 \hline
\end{tabular}
\caption{This table presents the value of each hyper-parameter used by BTAI in section \ref{ssec:BTAI_dSprites}.}
\label{tab:values_hp_BTAI_dSprites}
\end{table}

\vskip 0.2in
\bibliography{references}

\end{document}